\documentclass{article}

\RequirePackage{comment}
\RequirePackage{setspace}
\RequirePackage{float}
\usepackage{amsmath}
\usepackage{mathtools}
\DeclarePairedDelimiterX{\norm}[1]{\lVert}{\rVert}{#1}

\RequirePackage{algorithmicx}
\RequirePackage{algpseudocode}
\RequirePackage{graphics}
\RequirePackage{graphicx}
\RequirePackage{xcolor}

\RequirePackage[linesnumbered,ruled,vlined]{algorithm2e}
\SetKwInput{KwInput}{Input}                
\SetKwInput{KwOutput}{Output}              
\SetKwComment{Comment}{/* }{ */}

\usepackage{stackengine}

\usepackage{arxiv}
\usepackage[title]{appendix}

\usepackage[utf8]{inputenc} 
\usepackage[T1]{fontenc}    
\usepackage{hyperref}       
\usepackage{url}            
\usepackage{booktabs}       
\usepackage{amsfonts}       
\usepackage{nicefrac}       
\usepackage{microtype}      
\usepackage{cleveref}       
\usepackage{lipsum}         
\usepackage{graphicx}
\usepackage{natbib}
\usepackage{doi}

\usepackage{flushend}
\usepackage[percent]{overpic}
\usepackage{multirow}

\usepackage{xcolor}
\usepackage{caption}

\usepackage{float}
\usepackage{bm}
\usepackage{amssymb}

\usepackage{flushend}
\usepackage{stackengine}
\usepackage[percent]{overpic}
\usepackage{multirow}

\usepackage{xltabular}
\usepackage{lipsum}

\usepackage{acronym}
\newacro{GNC}{Guidance, Navigation and Control}
\newacro{ROS}{Robot Operating System}
\newacro{TEB}{Time Elastic Band}

\title{Blast Hole Seeking and Dipping – The Navigation and Perception Framework in a Mine Site Inspection Robot}

\newif\ifuniqueAffiliation
\uniqueAffiliationtrue

\ifuniqueAffiliation 
\author{ \href{https://orcid.org/0000-0001-5321-5396}{\includegraphics[scale=0.06]{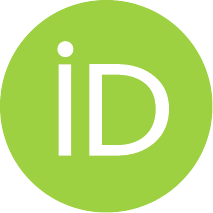}\hspace{1mm}Liyang~Liu}\thanks{Australian Centre for Robotics, USYD, at the time of this work being done. Contact email:  \texttt{Liyang.Liu@gmail.com}, \texttt{Andrew.Hill@sydney.edu.au}.} \\
	\And
\href{https://orcid.org/0000-0001-9516-0039}{\includegraphics[scale=0.06]{orcid.pdf}\hspace{1mm}Ehsan~Mihankhah} \\
\And
\href{https://orcid.org/0000-0001-6530-9138}{\includegraphics[scale=0.06]{orcid.pdf}\hspace{1mm}Nathan~Wallace} \\
\And
\href{https://orcid.org/0009-0008-9922-0164}{\includegraphics[scale=0.06]{orcid.pdf}\hspace{1mm}Javier~Martinez} \\
\And
\href{https://orcid.org/0000-0001-9297-4256}{\includegraphics[scale=0.06]{orcid.pdf}\hspace{1mm}Andrew~J.~Hill} \\
}
\else
\usepackage{authblk}

\author[1]{%
	\href{https://orcid.org/0000-0000-0000-0000}{\usebox{\orcid}\hspace{1mm}David S.~Hippocampus\thanks{\texttt{hippo@cs.cranberry-lemon.edu}}}%
}
\author[1,2]{%
	\href{https://orcid.org/0000-0000-0000-0000}{\usebox{\orcid}\hspace{1mm}Elias D.~Striatum\thanks{\texttt{stariate@ee.mount-sheikh.edu}}}%
}
\affil[1]{Department of Computer Science, Cranberry-Lemon University, Pittsburgh, PA 15213}
\affil[2]{Department of Electrical Engineering, Mount-Sheikh University, Santa Narimana, Levand}
\fi

\renewcommand{\shorttitle}{Blast hole Seeking and Dipping}

\hypersetup{
pdftitle={Blast hole Seeking and Dipping – The Navigation and Perception Framework in a Mine Site Inspection Robot},
pdfsubject={q-bio.NC, q-bio.QM},
pdfauthor={David S.~Hippocampus, Elias D.~Striatum},
pdfkeywords={First keyword, Second keyword, More},
}

\def\big{
	\begin{overpic}[height=6.8cm]{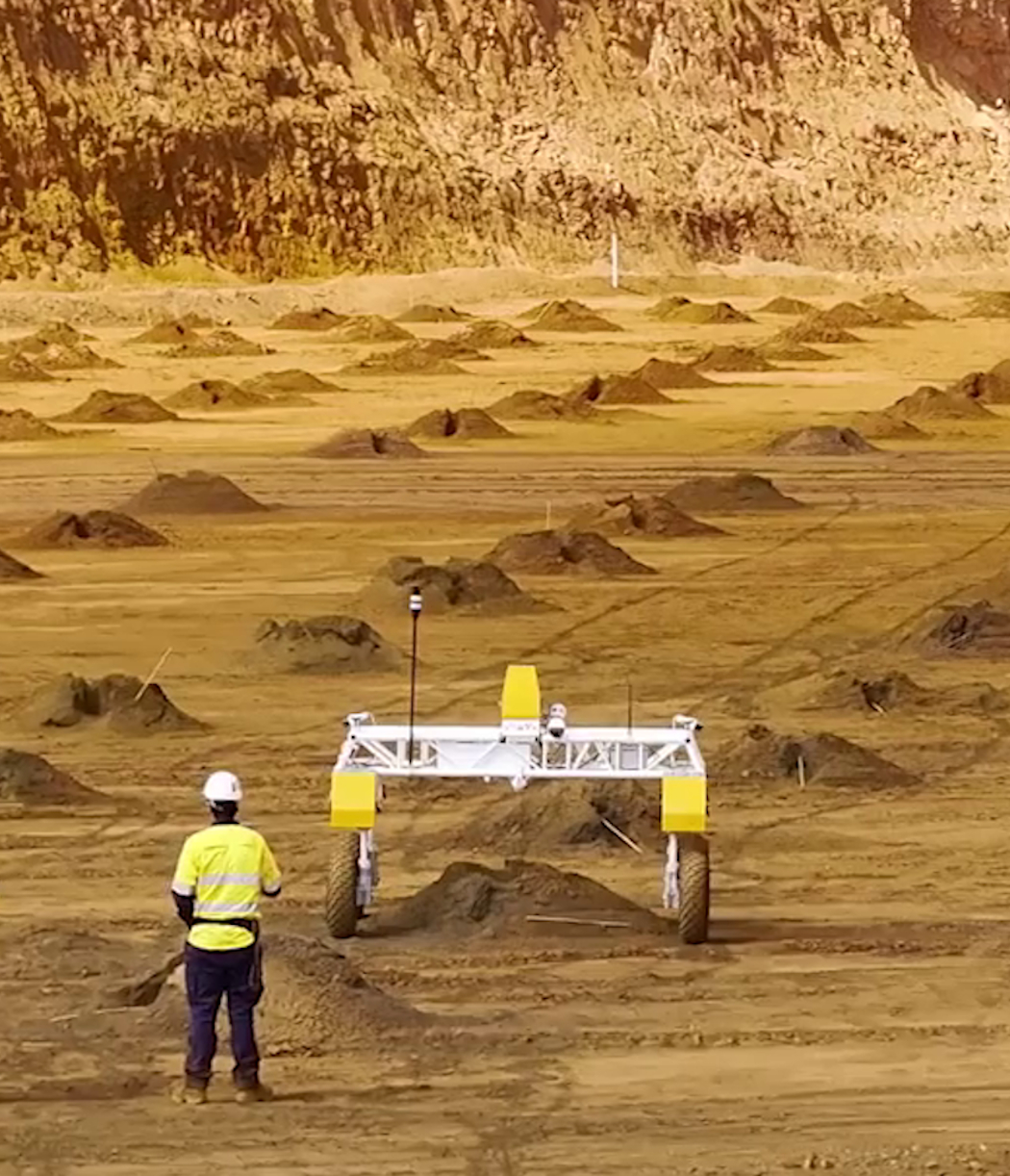}
		\put (2, 2) {\large{\color{red}{(a)}}}
\end{overpic}}
\def\little{
	\begin{overpic}[height=1.8cm]{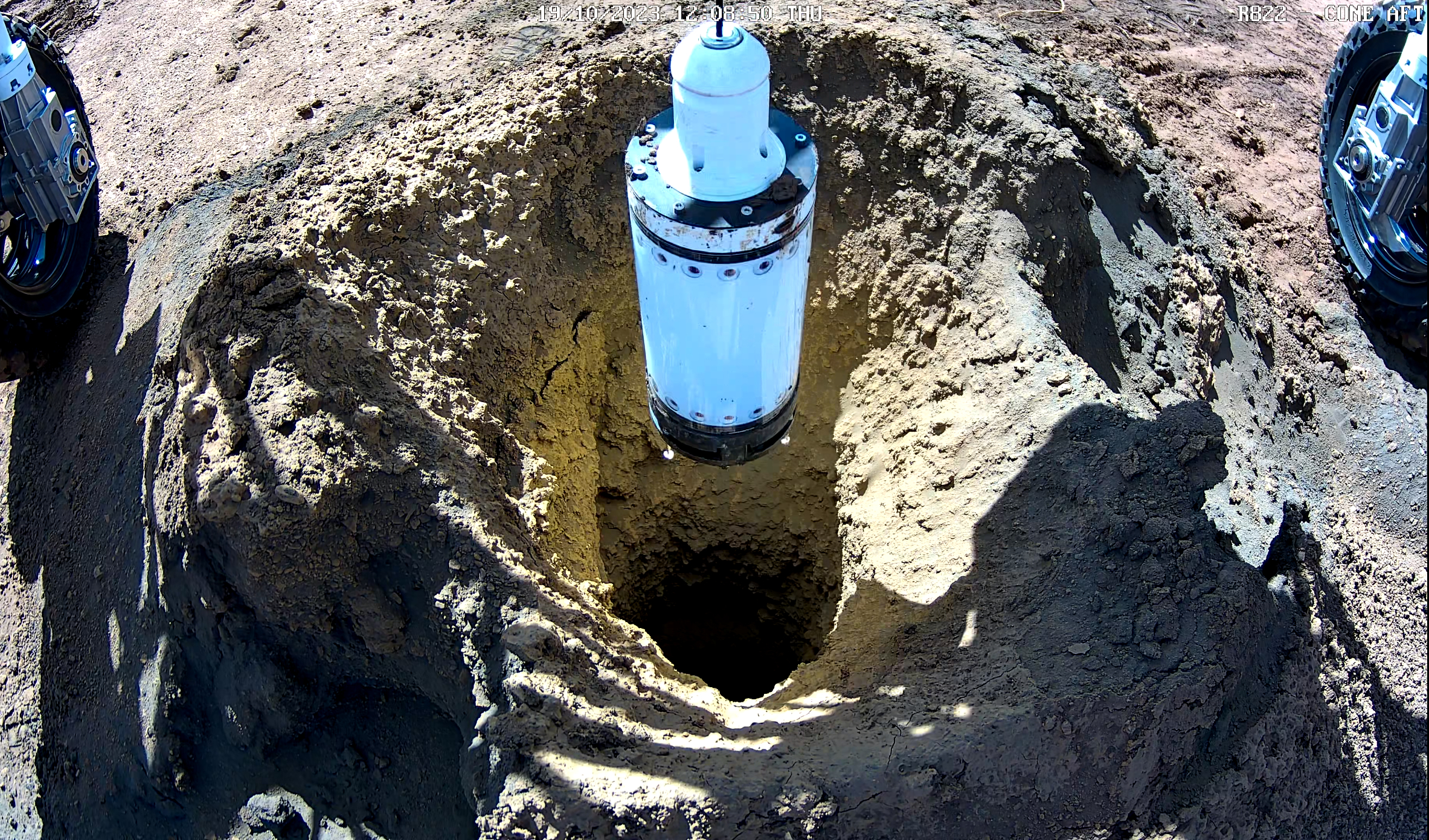}
		\put (6, 46) {\large{\color{red}{(b)}}}		
\end{overpic}} 

\begin{document}
\definecolor{DarkGreen}{rgb}{0.0,0.3,0.0},

\maketitle

\begin{abstract}
In open-pit mining, holes are drilled into the surface of the excavation site and detonated with explosives to facilitate digging. 
These blast holes need to be inspected internally to assess subsurface material types and drill quality, in order to significantly reduce downstream material handling costs. 
Manual hole inspection is slow and expensive, limited in its ability to capture the geometric and geological characteristics of holes.
This has been the motivation for the development of our autonomous mine-site inspection robot - "DIPPeR". 
In this paper, the automation aspect of the project is explained. We present a robust perception and navigation framework that provides streamlined blasthole seeking, tracking and accurate down-hole sensor positioning. 
To address challenges in the surface mining environment, where GPS and odometry data are noisy without RTK correction, we adopt a proximity-based adaptive navigation approach, enabling the vehicle to dynamically adjust its operations based on detected target availability and localisation accuracy. 
For perception, we process LiDAR data to extract the cone-shaped volume of drill-waste above ground, then project the 3D cone points into a virtual depth image to form  accurate 2D segmentation of hole regions. To ensure continuous target-tracking as the robot approaches the goal, our system automatically adjusts projection parameters to preserve consistent hole image appearance.
At the vicinity of the hole, we apply least squares circle fitting with non-maximum candidate suppression, to achieve accurate hole detection and collision-free down-hole sensor placement. 
We demonstrate the effectiveness of our navigation and perception system in both high-fidelity simulation environments and on-site field trials.
A demonstration video is available at \url{https://www.youtube.com/watch?v=fRNbcBcaSqE}.
\end{abstract}


\section{Introduction}
\label{sec:intro}
In mining, blast holes are vertical holes drilled into the surface of the excavation site. They are packed with explosive material, and detonated to induce cracks in the surrounding rock to facilitate digging. 
Holes may be inspected for quality assurance, operational and geological reasons \cite{Leung-surface-mining-magzine}.
The blast hole site, however, can be hostile to human operators due to safety hazards and
extreme weather conditions. This motivates the development of the Downhole Inspection, Probing, and Perception Robot (DIPPeR) for mine-site inspection, a research task carried out at Rio Tinto Sydney Innovation Hub (RTSIH), The University of Sydney.

While robotic research has made significant progress in recent years, 
the key challenge of applying robots to industrial problems remains the need for robustness, repeatability, and cost-effectiveness \cite{darpa-data-61}. Practical implementation often requires finding balance between these factors \cite{pepper-harvest-robot}, and this has driven our design decision to implement a proximity-based adaptive navigation approach; tailoring operations to the changes in availability and accuracy of the robot's observations. We address this through consideration of the navigation, sensing, and data availability challenges discussed in the following sections.

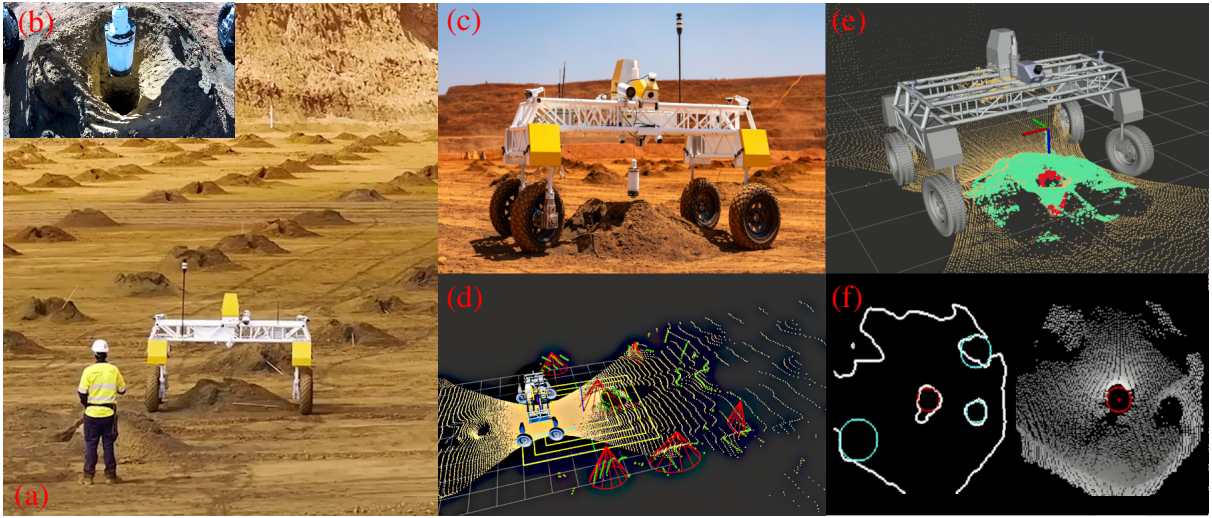
\begin{figure*}[!t]	
		\centering
		\begin{tabular}{@{}l@{}l@{}}		
			\stackinset{l}{0pt}{t}{0pt}{\little}{\big}
			&
			\begin{overpic}[width=0.62\linewidth]{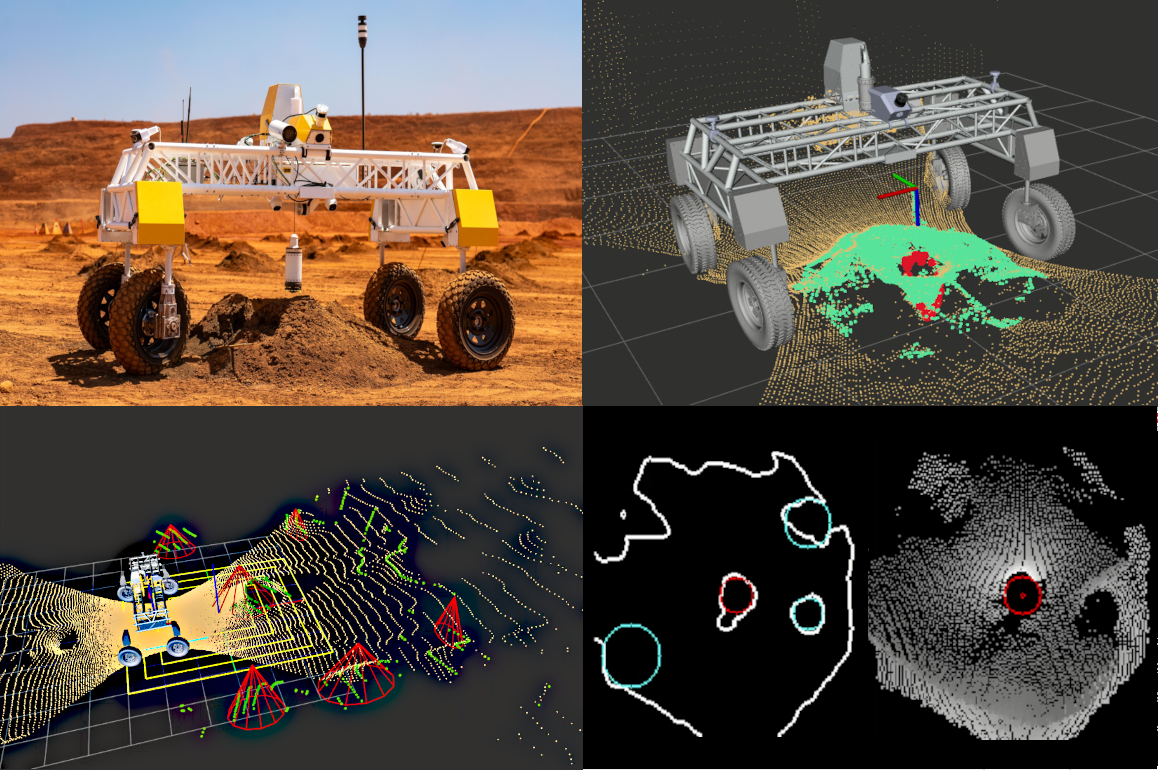}
				\put (1,63) {\large{\color{red}{(c)}}}
				\put (1,27) {\large{\color{red}{(d)}}}
				\put (51,63) {\large{\color{red}{(e)}}}
				\put (51,27) {\large{\color{red}{(f)}}}
			\end{overpic}\\
		\end{tabular} \\
		\captionof{figure}{ (a) DIPPeR robot inspects blast holes in regular pattern in open-pit mine. (b) and (c) Sensor sonde dipped autonomously into blast hole.  (d) DIPPeR uses LiDARs to scan environment: it detects ground plane (bright yellow), above ground points (green) and cones (red pyramids).  (f) Cone point cloud projected into 2D. Hole-detection candidates identified (cyan) and maximum aposterior selection (red). \\
			Note for Subfigures (a,b,c): \textbf{ \textcopyright{Rio Tinto} 2025, All Rights Reserved.} }			
		\label{fig:hole-teaser}
\end{figure*}


\textbf{Control and Planning} \\
There are many approaches to solve an automation task. From a robot-theoretic point of view, the common practice ~\cite{2022-fr-eiffert-autonomy}~\cite{2023-vision-nav-review}~\cite{2020-mbz-mobile-manip-wall} would be to first explore the environment to generate a map, then carefully plan efficient paths to navigate between objects (in this case blastholes within a blasthole pattern). Although this strategy offers the prospect of full autonomy in traversing diverse environments, 
developing such an all-purpose solution for mine-site inspection would incur high cost and produce unnecessary features. Due to the transient nature of the blast sites---with short times between drilling and blasting---and destructive nature of the operation, mapping is of negligible value and the robot focus should be on hole inspection.


Therefore we choose to use the prior knowledge about the fairly structured environment, demonstrating that this is sufficient for the task and introduces minimal overhead. First it is guaranteed that Blast-site GPS is reliable and ubiquitous. Secondly, Blast benches are well-structured, with blast holes bored in a regular grid pattern on the flat bench surface. Finally the hole GPS positions  were pre-recorded during drilling for convenience of post-drilling revisits. This enables the DIPPeR robot to utilise its on-board GPS receiver to navigate to each hole to a coarse accuracy, bringing the blast hole cone in detection range before switching to a more accurate relative localisation mode for proximity operations, thereby eliminating the need for mapping operations. Further, since the robot motion will be confined to straight lines along hole columns, an in-place turn would allow the robot to steer to arbitrary directions and transfer between columns. The seeking and dipping cycle can be repeated to complete the whole site inspection task.

\textbf{Sensing and Perception} \\
Since the geo-referencing of the holes is insufficiently precise for reliable positioning of the sensor probe, suitable sensor selection, accurate target detection and tracking become crucial.

Although monocular cameras as a vision sensor have been reported to successfully detect blast hole profiles in \cite{bhole-svm}, this is in the context of air-borne surveying for statistical hole data collection with profile-level accuracy. For close-distance applications, the typical colors seen at the mine-site are dependent on location and orebody;  uniform color and lack of texture is common. Due to low mounting heights, factors such as extreme illumination conditions, self-shadowing, insufficient field of view affect image quality, making cameras difficult and unreliable to use. 


However, the LiDAR sensor provides more resilience to changing illumination and a wider field of view, and is known to be reliable in mine-sites~\cite{surface-mining-sensor}. As shown in Figure \ref{fig:hole-teaser}, the blast-site is maintained as a flat plane. Each blast hole is surrounded by a cone-shaped drill-waste pile, referred to as the cone in this paper. A LiDAR scan can reveal the cone above the ground. Further, the opening of the hole does not have LiDAR returns due to the absence of material; this is visually apparent when viewed from above. This becomes the single most reliable feature for hole detection and tracking. 

Using an approach similar to \cite{rivetDetect}~\cite{rectangular}, we project the 3D LiDAR points downward into a virtual depth image then perform hole detection in the image domain.  Exploiting the fact that no LiDAR return beams come from the hole interior, our depth image is naturally segmented with an apparent void in the centre bounded by the cone collar (Figure \ref{fig:hole-teaser}(f)); a prominent feature for visual servoing.

Once detected, the holes need to be tracked throughout the robot's approach. In conventional machine vision, the camera moves with the robot, and visual tracking involves searching for similar pixels in adjacent camera frames, using methods such as Optical Flow \cite{Vins-mono} or feature tracking \cite{orbslam-TRO2015}. In our system the virtual camera remains fixated above the cone, separate from the moving robot. Tracking in this case refers to maintaining consistent hole detection and positioning of the associated virtual camera. The hole's non-uniform shape, together with variation in the LiDAR coverage, density, and extent of hole occlusion at various distances and incidence angles, can cause significant changes in the apparent position and size of the hole in the image frame.

Phantom holes due to irregular cone shape and LiDAR occlusion can appear and lead the robot astray, we apply methods in both 3D and 2D domains to minimise phantom holes' effect.

Once straddling the cone, the sonde-to-hole alignment becomes critical for mission success. 
DIPPeR is expected to visit blast holes with diameters ranging from 24cm to 30cm.
This in some cases only permits a few centimetres clearance for the sonde unit, thus requiring precise relative positioning for successful inspection. To maximize processing effects, accurate handling takes place in the image domain. The authors of \cite{bhole-svm} demonstrated how to extract Histogram of Oriented Gradients (HOG) features~\cite{hog_1986} from the images in a 2D linear  processing order. By feeding these features into the Support Vector Machine (SVM), they were able to achieve profile-level accuracy in detecting hole regions. We will show in this paper that in our projected virtual images, circular features are a better choice than translational features owing to its continuity and smoothness, and we can achieve highly accurate hole localisation for in-situ dipping and downhole inspection.


\textbf{Data Availability}\\
Due to safety regulations there is often no opportunity for a-priori data collection on the mine site, and provided information is mainly the mean bore hole diameter and an approximate range of cone heights.
In \cite{pothole-mehala} ground height map was exploited for learning-based pothole detection, this does not apply to our meter-grade deep holes. 
Further manual labelling for data-driven detection is costly and impractical. We propose a cone/hole detection pipeline which leverages classical computer-vision techniques, avoiding the use of data-hungry learning methods. This results in a processing pipeline which is simple to configure and tune to the requirements of each site without requiring extensive data collection, labelling and training.

In this paper, we present the LiDAR based blasthole detection system used in the DIPPeR robot hole seeking process. We first give an overview of our inspection robot system in Section \ref{sec:system-overview}.  Next, we present our navigation strategy \ref{sec:nav-overview}. Then in Section \ref{sec:detect-track} we show how to detect the cone target and how to track the target by auto-adjusting the perceptive parameters during robot motion. In Section \ref{sec:detect-hole}, we show first how to form well-segmented binary images with well connected white cone face and black hole, then we present our two-stage coarse to fine detection process that achieves an excellent detection accuracy by smart region extraction, optimal circle fitting, and non-maximum-suppression. 
In the experiment section \ref{sec:experiment} we give thorough trial test results.

\section{System Overview}
\label{sec:system-overview}
\begin{figure*}[!t]
	\centering
	\includegraphics[width=0.95\textwidth]{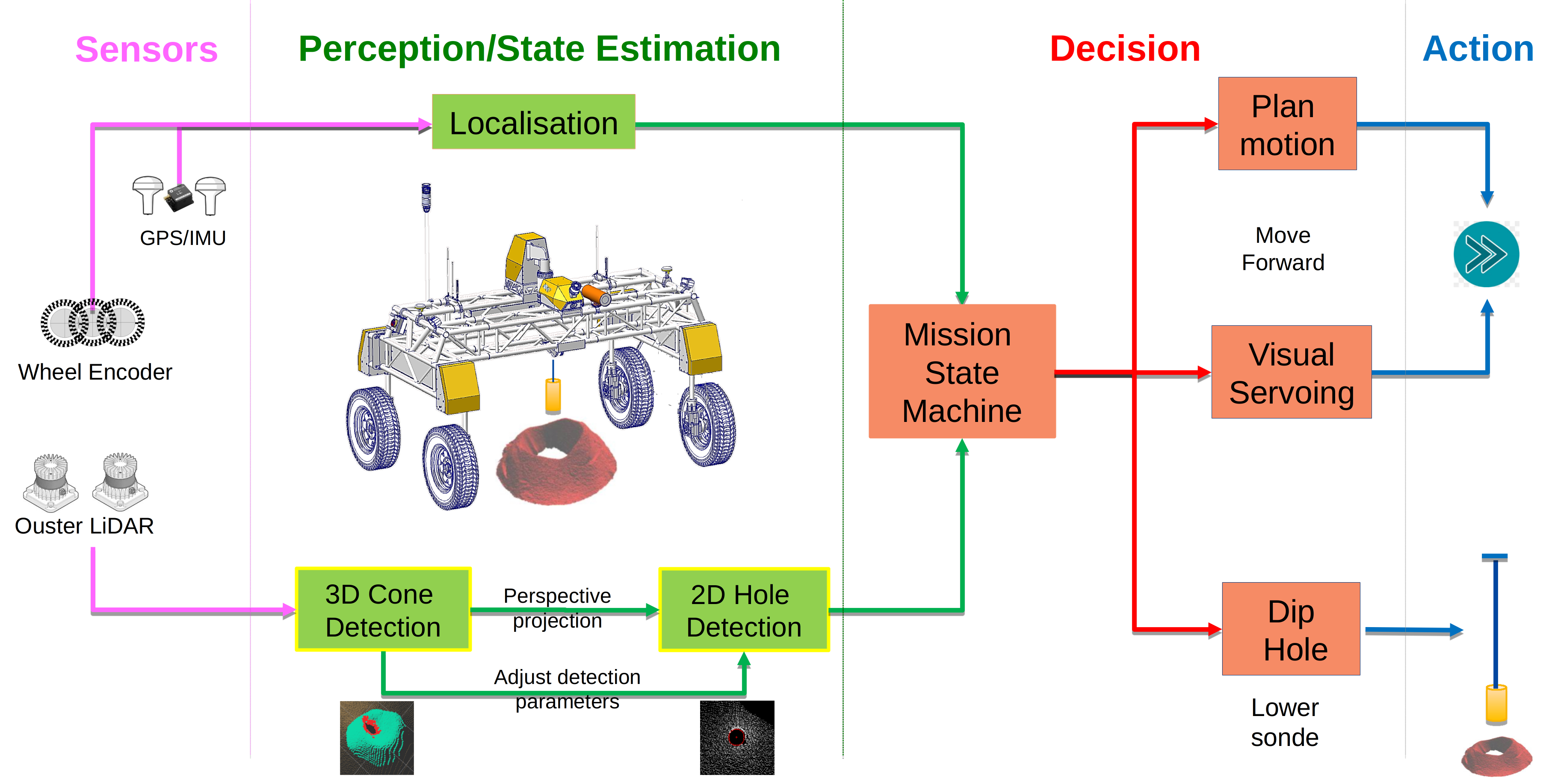}
	\caption{Overview of the core system modules and control flow.}
	\label{fig:overview}
\end{figure*}

The main task for DIPPeR is to visit and inspect each blast hole in a minesite inspection mission. As illustrated in Figure \ref{fig:overview}, the sensing and perception layers of the robotic system consists of following components: sensors, perception and state estimation, mission manager, motion planner, base controller and dipping actuator.

The GPS system continuously outputs robot poses in the global UTM coordinates. The localisation module outputs robot pose in the local coordinates frame (``odom'', origin at start of mission), this is estimated from the proprioceptive sensors. Robot poses from both coordinate frames are fed to the decision making module -- the mission manager. The perception module is another continuously running sub-system, reporting detected cones and hole positions to the manager whenever they are available.
The mission manager maintains a state machine, which continuously compares DIPPeR's position in  all coordinate frames, against either the designated waypoint GPS positions or detected target positions. Two possible actions come from manager: navigate to the target or dip the hole.

Depending on the distance between the robot and the cone, as well as the cone's height, the perception module auto-adjusts its target detection settings. This ensures consistent tracking of the target hole as the robot approaches.

\section{Navigation Strategy}
\label{sec:nav-overview}

Our mission controller is a type of cyclic state state machine as depicted in Figure \ref{fig:state-machine}(b). Our proxmity-based adaptive navigation strategy is the backbone of this state machine, where each state follows its own coordinate frame and localisation method, and is completely decoupled from one another.

\subsection{Decoupled Localisation}
\label{sec:coords}
\begin{figure}[h]
	\centering
	\setlength\tabcolsep{3pt}	
	\begin{tabular}{cc}
		{%
			\setlength{\fboxsep}{2pt}%
			\setlength{\fboxrule}{1pt}%
			\fbox{\includegraphics[width=0.35\linewidth,height=5.5cm]{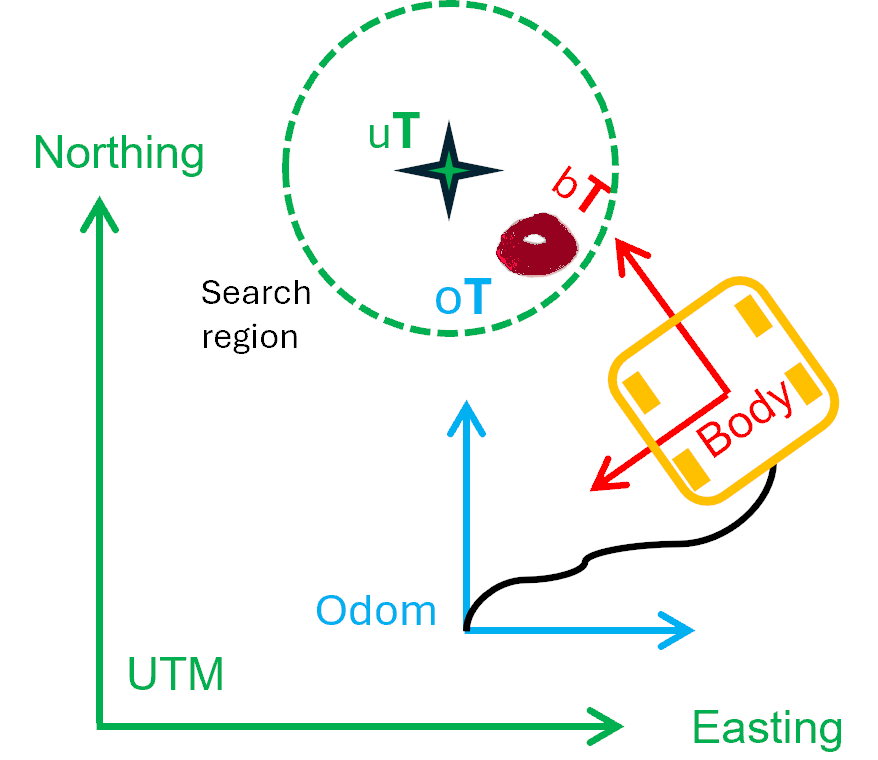}}
		}
		&
		{		\setlength{\fboxsep}{2pt}%
			\setlength{\fboxrule}{1pt}%
			\fbox{\includegraphics[width=0.58\linewidth, height=5.5cm]{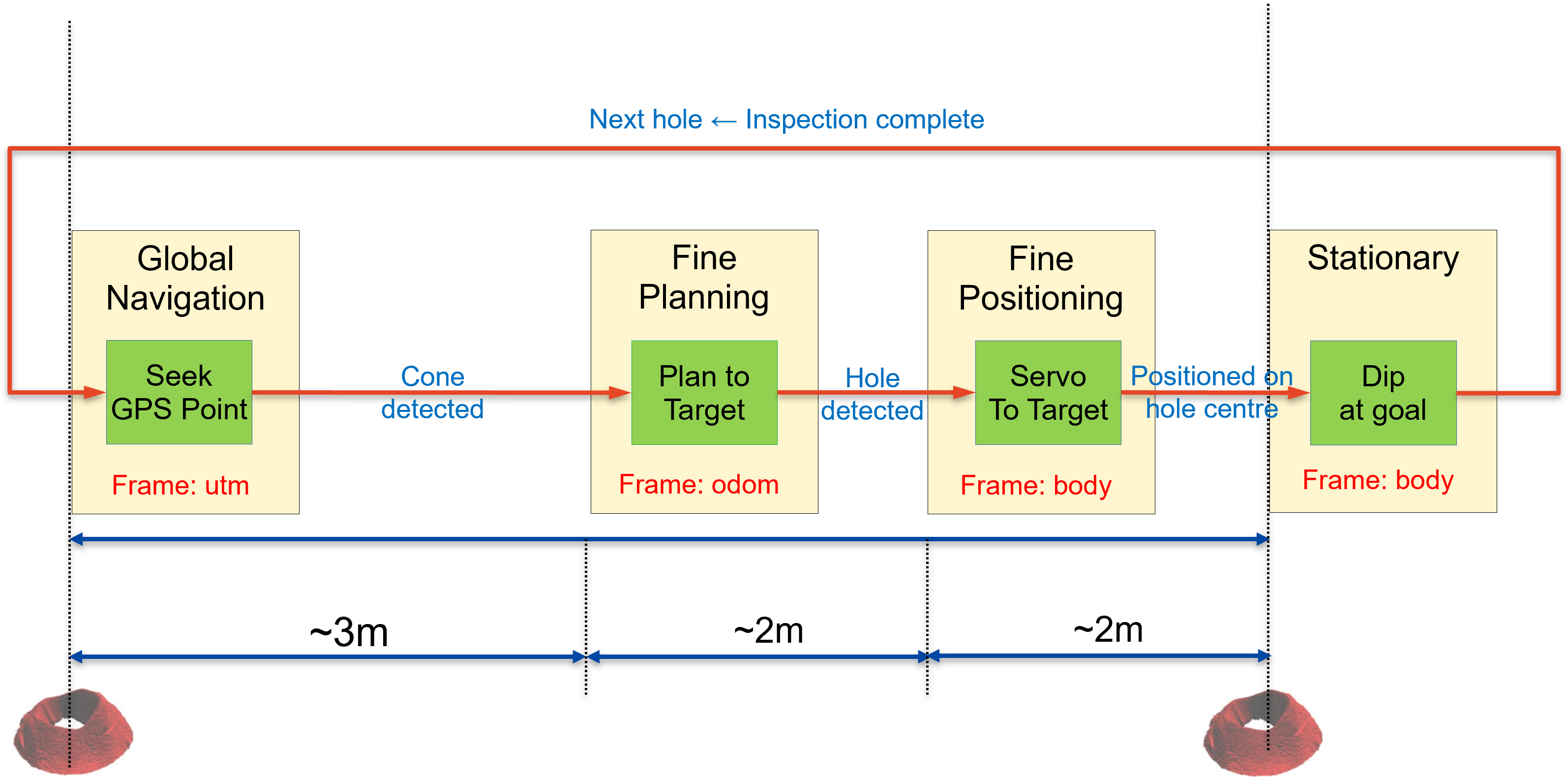}}
		} \\	
		\begin{tabular}{p{0.35\linewidth}}
			(a) DIPPeR maintains three separate ccoordinate frames for localisation:
			UTM(\textcolor{DarkGreen}{green}), Odom(\textcolor{blue}{blue}), and Body(\textcolor{red}{red}).\\ 
		\end{tabular}
		& 
		\begin{tabular}{p{0.55\linewidth}}
			\small{(b) Mission state machine: (i) seek the given GPS point to coarse accuracy, (ii) after detecting the cone do fine motion planning, (iii) after detecting the hole do fine visual servoing, (iv) after good alignment dip the hole.}
		\end{tabular}
	\end{tabular}	
	\caption{The seek-hole navigation cycle relies on decoupled localisation: use the \textcolor{DarkGreen}{UTM} frame for GPS-point seek and regional target search, \textcolor{blue}{Odom} for planning to target, and \textcolor{red}{Body} for fine servoing, totally avoiding GPS and odometry fusion.}
	\label{fig:state-machine}
\end{figure}

As shown in Figure \ref{fig:state-machine}(a), there are three coordinate conventions used in this navigation system, namely: the universal transverse mercator (UTM) reference frame, the local Odom frame where the origin is at inspection start position, and the robot Body frame. Both the UTM and Odom frame following the Easting-Northing-Up(ENU) convention. And the robot body frame follows a Forward-Left-Up (FLU) convention. 
We use the symbol  ${}^{\mathrm{F}}\mathbf{T}$ to denote the target position, in a reference frame ${}^\mathrm{F}$ in superscript font on the left.

Each hole to be inspected has a  pre-recorded GPS position ${}^{\mathrm{U}}\mathbf{T}$ in the UTM frame from the driller. This term has a meter-grade error from the actual position. We define a circular search region centred at the designated point, with its radius set to 4 meters to sufficiently cover the GPS signal error in absence of RTK correction. Any cone detected in this region is considered a possible candidate to contain the actual hole. In the event of multiple detection, the cone with the smallest Line-of-Sight (LOS) angle to the robot heading is chosen as the matching hole for inspection. 

When the robot initially detects a target hole in its body frame, denoted ${}^{\mathrm{B}}\mathbf{T}$, by body to global transformation, the target is locked to ${}^{\mathrm{U}}\mathbf{T}$ in the "UTM" frame. Similarly, by body to world transformation, the target can be locked to ${}^{\mathrm{O}}\mathbf{T}$ in the "Odom" frame. 

As DIPPeR moves around, and in absence of RTK correction, its GPS measured position ${}^{\mathrm{U}}\mathbf{D}(t)$ in the UTM frame exhibits frequent jumps due to factors like atmospheric disturbances and satellite orbital variances. Its distance to a detected target in this frame ${}^{\mathrm{U}}l=|{}^{\mathrm{U}}\mathbf{D}(t) - {}^{\mathrm{U}}\mathbf{T}|$ also exhibits corresponding shifts in value. UTM position guided navigation works well for long distance travel where the accuracy requirement is low. However, for close-distance navigation, a controller that relies on the erroneous UTM distance would lead to random steering behaviour, losing sight of the target to eventual navigation failure. 

On the other hand,  the robot's position in the Odom frame  ${}^{\mathrm{O}}\mathbf{D}(t)$ evolves smoothly as it is computed from continuous integration of odometer measurements, and is an accurate estimate of the robot pose over a short duration. The "Odom" frame target distance ${}^{\mathrm{O}}l=|{}^{\mathrm{O}}\mathbf{D}(t) - {}^{\mathrm{O}}\mathbf{T}|$ follows a steady decreasing trend to settlement. A controller that relies on this Odom distance should lead to eventual target contact. This is the motivation behind our decoupled localisation scheme, no attempt is made at fusing the GPS measurements with local odometry during the entire inspection mission.

\subsection{Control Paradigm}
Upon mission start, the robot navigates towards the GPS position of the first blast hole in the sequence, relying on its on-board GNSS antenna for position feedback, and navigating in the UTM frame.


Once within $\sim$4 metres of the blast hole cone, DIPPeR can reliably detect the target, thus triggering the fine motion planning stage. The robot plans its trajectory with the detection result as the goal, moving to track the target continuously as the distance shortens. Localisation is performed in the local "Odom" frame to produce smooth pose estimates and avoid any discontinuities due to GPS drift or error. 
Further, the robot may re-align its heading to the cone centre during this stage if the LOS exceeds a certain threshold.

Once in a meter proximity to the target, the robot enters the fine positioning (visual-servoing) stage. At each step, a velocity command is issued in the robot body frame based on current relative target pose. Servoing on the detected hole centre continues until the sonde is precisely aligned.
This process repeats for each of the holes in the sequence, negotiating a boustrophedon coverage pattern \cite{choset1998coverage} over the grid of blast holes.


The proposed strategy fully utilises the known structure of the blast bench in its planning and execution, eliminating costly mapping operations and mitigating the impacts of GPS localisation discontinuity.
\vspace*{5mm}

\section{Target Detection and Tracking}
\label{sec:detect-track}

As shown in Figures \ref{fig:hole-teaser} and \ref{fig:hole-over-view}, 
cones and holes are detected by processing LiDAR point clouds and projected images. As the robot moves forward, the cone target is first detected when the LiDAR scan reveals the front surface of an above ground object. As the robot nears, the hole at the cone centre is gradually resolved, which then becomes a more reliable tracking target for the subsequent servoing operations.


\subsection{Ground Tilt Correction}
\label{sec:tilt-correction}

The first step in LiDAR perception is about transforming the point clouds to a ground aligned coordinates frame. This allows simple blast cone points extraction by thresholding on points height. 

In instances where the robot straddles
large cones and experiences a tilt relative to the ground, the
on-board orientation sensor can be used to apply a correction.

To this end, we make use of the on-board IMU data. The IMU reading includes the robot's absolute position and orientation in the geo-referenced frame.  By analysing the relative rotation between the ground plane normal $\hat{\mathbf{n}}_G$ with the robot base normal $\hat{\mathbf{n}}_R$, we can correct for the roll-pitch tilt experienced in the robot.

\begin{figure}[h]
	\centering
		\setlength{\tabcolsep}{3pt}
		\begin{tabular}{cc}
			\includegraphics[width=0.44\linewidth,height=5.3cm]{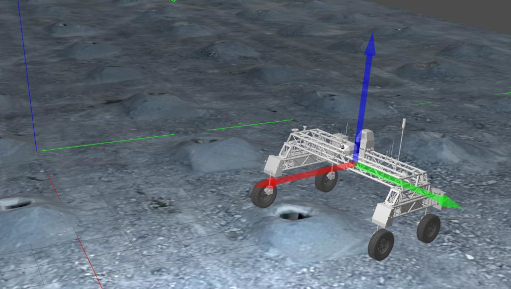}
			&
			{%
				\setlength{\fboxsep}{0pt}%
				\setlength{\fboxrule}{2pt}%
				\fbox{\includegraphics[width=0.44\linewidth, height=5.2cm]{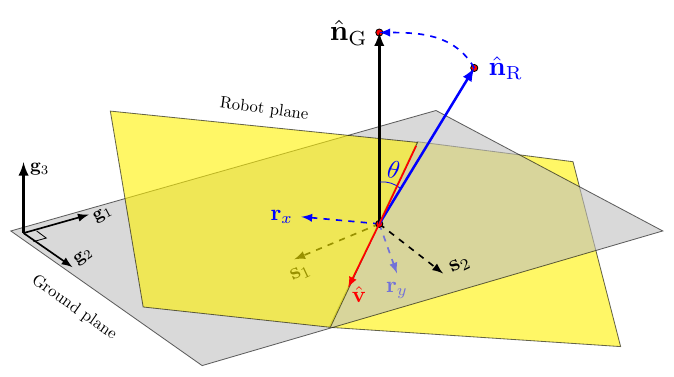}}
			} \\
			\begin{tabular}{p{0.41\linewidth} }
				(a) Tilted robot in simulation.\\ \textbf{\textcopyright Rio Tinto 2025, All Rights Reserved. 3D surface model generated using  DroneDeploy, Inc. software. Visualised using  Gazebo $[$11.0/Open-Robotics$]$}.
			\end{tabular}
			&
			\begin{tabular}{p{0.41\linewidth} }
				(b) Robot frame correction. Correcting robot tilt is equivalent to rotating the base plane about vector $\hat{\mathbf{v}}$ by angle $\theta$ to re-align the vehicle's X-Y axis with their projections on the ground.
			\end{tabular}
		\end{tabular}
	\caption{Robot base tilt correction}
	\label{fig:base-tilt}
\end{figure}
We make use of the robot's shadow (approximately) on the ground to define a coordinate frame level to the ground at the robot's position. LiDAR point clouds should be transformed to this shadow's frame for further processing. For that, first we find the rotation in the geo-referenced frame that aligns the robot base to its shadow.


\begin{equation}
\begin{aligned}
\text{Given }\qquad&\text{true ground normal :} 
\\
\hat{\mathbf{n}}_{\mathrm{G}} &= [\, x_{\mathrm{G}}, \, y_{\mathrm{G}}, \, z_{\mathrm{G}} \,]^\intercal \approx [0, 0, 1]\, ,\\
\text{and }\qquad&\text{robot orientation in geo-frame}\\ \mathbf{R}_{\mathrm{geo}} &= [\mathbf{r}_x, \mathbf{r}_y, \mathbf{r}_z]:
\\
\text{so }\qquad&\text{robot base normal } \hat{\mathbf{n}}_\mathrm{R} \text{ in geo-frame } :
\\
\hat{\mathbf{n}}_{\mathrm{R}} &= \mathbf{r}_z = [\, x_{\mathrm{R}}, \, y_{\mathrm{R}}, \, z_{\mathrm{R}} \, ]^\intercal.
\end{aligned}
\end{equation}

We derive the rotation matrix (in the geo-frame) to align the robot's base plane to its shadow on the ground. We can do so using cross product and the exponential map \cite{2018-wafr-pmba}.
\begin{equation}
\begin{aligned}
\mathbf{R}_{\mathrm{align}} &= \mathrm{Exp}(\, \hat{\mathbf{v}} \, \theta \, ) \, ,
\\
\hat{\mathbf{n}}_\mathrm{G} &= \mathrm{R}_{\mathrm{align}} \, \hat{\mathbf{n}}_{\mathrm{R}} \, ; 	
\\
\\
\text{where }\quad 
\hat{\mathbf{v}} &=  \frac{\hat{\mathbf{n}}_\mathrm{R} \times  \hat{\mathbf{n}}_\mathrm{G} }
{\|\hat{\mathbf{n}}_\mathrm{R} \times \hat{\mathbf{n}}_\mathrm{G}\| } \, , 
\\
\quad \theta &= \mathrm{atan2}(
\,\,
\|\hat{\mathbf{n}}_\mathrm{R} \times \hat{\mathbf{n}}_\mathrm{G}\|,
\,\,
\hat{\mathbf{n}}_\mathrm{R} \cdot \hat{\mathbf{n}}_\mathrm{G}
\,\,
) \, .
\\
\end{aligned}
\end{equation}

\noindent Now we can work out orientation of the robot's shadow in the absolute geo-frame.
\begin{equation}
\mathbf{R}_{\mathrm{shadow}} = \mathbf{R}_{\mathrm{align}} \mathbf{R}_\mathrm{{geo}} 
\end{equation}
The axis of rotation in the shadow's frame is
\begin{equation}
\hat{\mathbf{v}}^{\mathrm{S}} = \mathbf{R}_{\mathrm{shadow}}^\intercal \hat{\mathbf{v}}
\end{equation}

\noindent Therefore the robot orientation in its shadow frame is:
\begin{equation}
\mathbf{R}_{\mathrm{R}}^{\mathrm{S}} = \mathrm{Exp}(\, -\hat{\mathbf{v}}^{\mathrm{S}} \, \theta \, ) 
\end{equation}
This should be the rotation to transform LiDAR points from the robot frame to the ground aligned frame.


\subsection{Cone Detection}
\label{sec:cone-detect}
Against the flat ground of the bench, cones can be detected by height thresholding. We use long-range low resolution LiDAR for far distance cone detection. Once the robot gets close to 3 meter distance,  DIPPeR switches to the short-range high resolution LiDAR for cone and hole detection. Further, a series of box filters are applied to eliminate LiDAR points  reflected off the robot body parts, including the four wheels and the chasis box.

Using the assumption that the bench is free from foreign objects, we classify any large point cluster contained within the 6-meter corridor in front of the robot and above a certain size as the cone object, avoiding computationally-expensive 3D shape analysis. After removing flying noise points hovering above the cone surface  or inside the hole cavity, the filtered cloud looks like a clean surface.

The cone centre needs to be computed and will be used in subsequent depth image projection. To avoid biasing due to uneven distribution of LiDAR points, we use a 2D voxel-grid to group the cone points, then compute a weighted average of the non-empty voxels, using voxel height as the weight, to form a fair centroid estimate. 

Due to frequent geological sampling, the cone surface may contain shallow pits around its edge (Figures \ref{fig:hole-teaser}(a) and \ref{fig:notch}). These pits can be erroneously detected as the blastholes causing the robot to go astray. To eliminate these `phantom' holes, we fit a Convex Hull to retain LiDAR points from its shallow base that lie within the cone circumference. This ensures the 2D projected cone face appears consistent and unmarred by pitting. Further phantom hole treatment will be applied in the 2D image domain in Section \ref{sec:n-m-s}.

\subsection{Virtual Image Formation}

\begin{figure*}[!t]
		\centering
		\begin{tabular}{@{}c@{}}
			\includegraphics[width=0.92\linewidth]{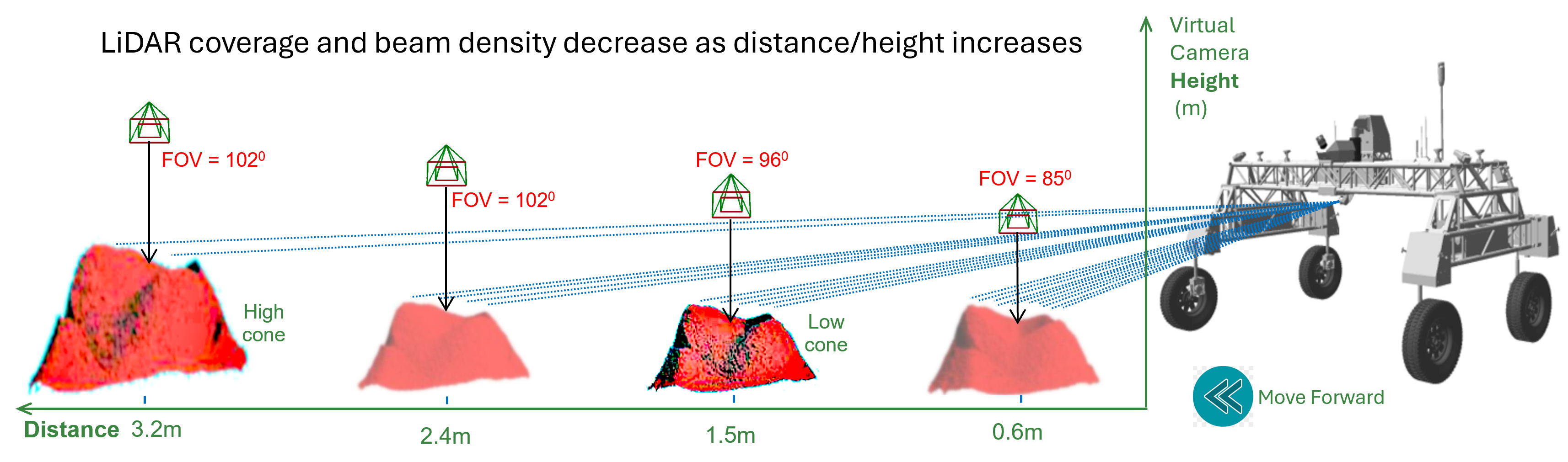} \\
			\\ [-18pt]
			\setlength{\tabcolsep}{1pt}
			\begin{tabular}{ccccc} 
				\includegraphics[width=0.18\linewidth]{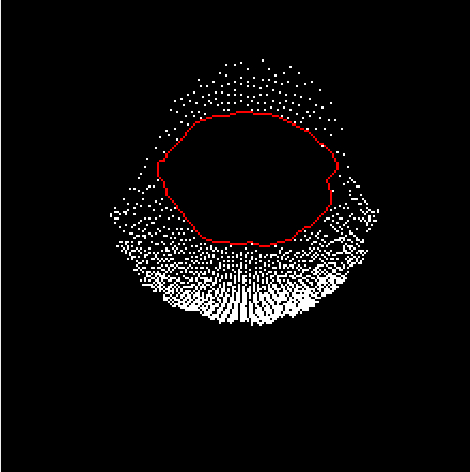} 
				&
				\includegraphics[width=0.18\linewidth]{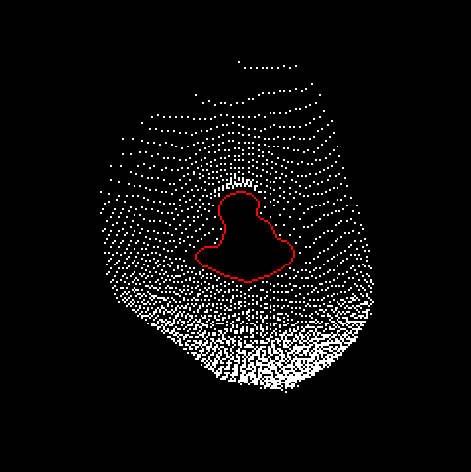}
				&		
				\includegraphics[width=0.18\linewidth]{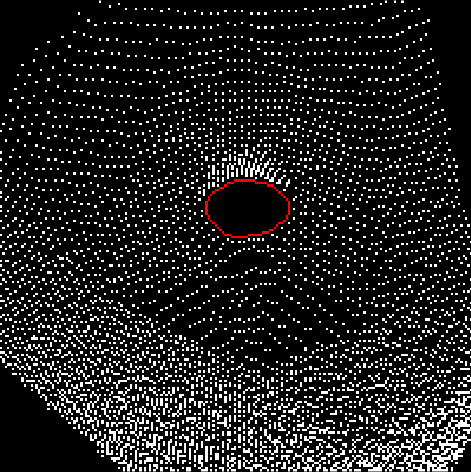}		
				&
				\includegraphics[width=0.18\linewidth]{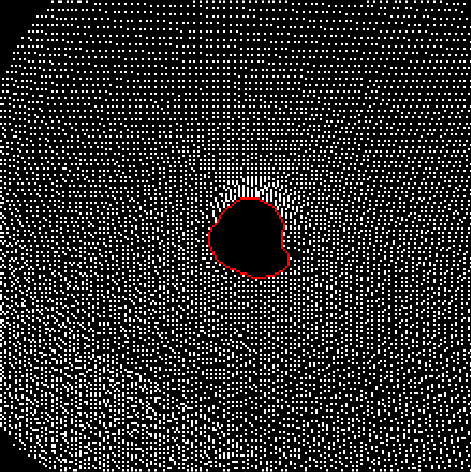}		
				&
				\includegraphics[width=0.18\linewidth]{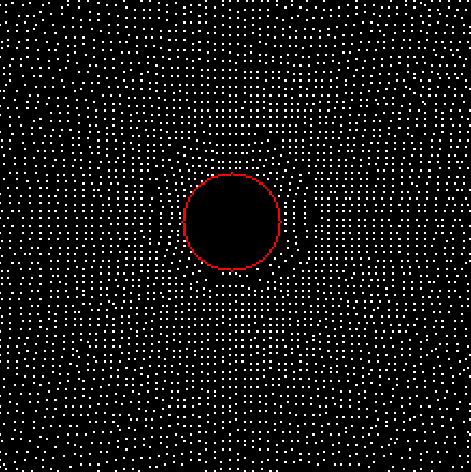}		
				\\
				\small{(a) Coarse-projection}
				& 			\small{(b) Coarse-projection}
				& 			\small{(c) Coarse-projection}
				& 			\small{(d) Coarse-projection}		
				& 			\small{(e) Fine-projection}
				\\
				\small{3.2(m) / 2.5(m) / $102^\circ$}
				&
				\small{2.2(m) / 2.2(m) / $102^\circ$}
				&
				\small{1.6(m) / 1.8(m) / $96^\circ$}
				&
				\small{0.6(m) / 1.6(m) / $84^\circ$}
				&
				\small{0.2(m) / 1.3(m) / $71^\circ$}
				\\
			\end{tabular}\\
		\end{tabular}
		\captionof{figure}{At various distances $D$ to the robot, by adjusting the virtual camera's height $H$ and Field-of-View $FoV$: the hole is consistently located in the centre of the virtual images with similar sizes. The large hole opening of high cone is due to hole's funnel shape and beam occlusion. Projection setting include ``Distance/Height/FoV''}		
		\label{fig:dist-vs-camera}
\end{figure*}

Hole detection takes place in 2D image domain. These images are generated by projecting the cone cloud through a virtual camera at different positions above the cone. Based on projective geometry principle \cite{Hartley_Zisserman_2004}, camera calibration settings such as camera height and focal length (aka field of view) affect the virtual image in the following aspects: cone face coverage, pixel spacing, and hole object size. This is illustrated in Figure \ref{fig:dist-vs-camera}. 
Further, non-central projection causes the hole to appear more elliptical. 	To address this, the virtual camera settings should be auto-adjusted at various distances during target tracking.

\subsection{Tracking with Distance dependent Image Projection}
\label{sec:dist-range}

\begin{figure*}[!t]
	\centering
      \begin{tabular}{cc}
			\includegraphics[width=0.60\linewidth]{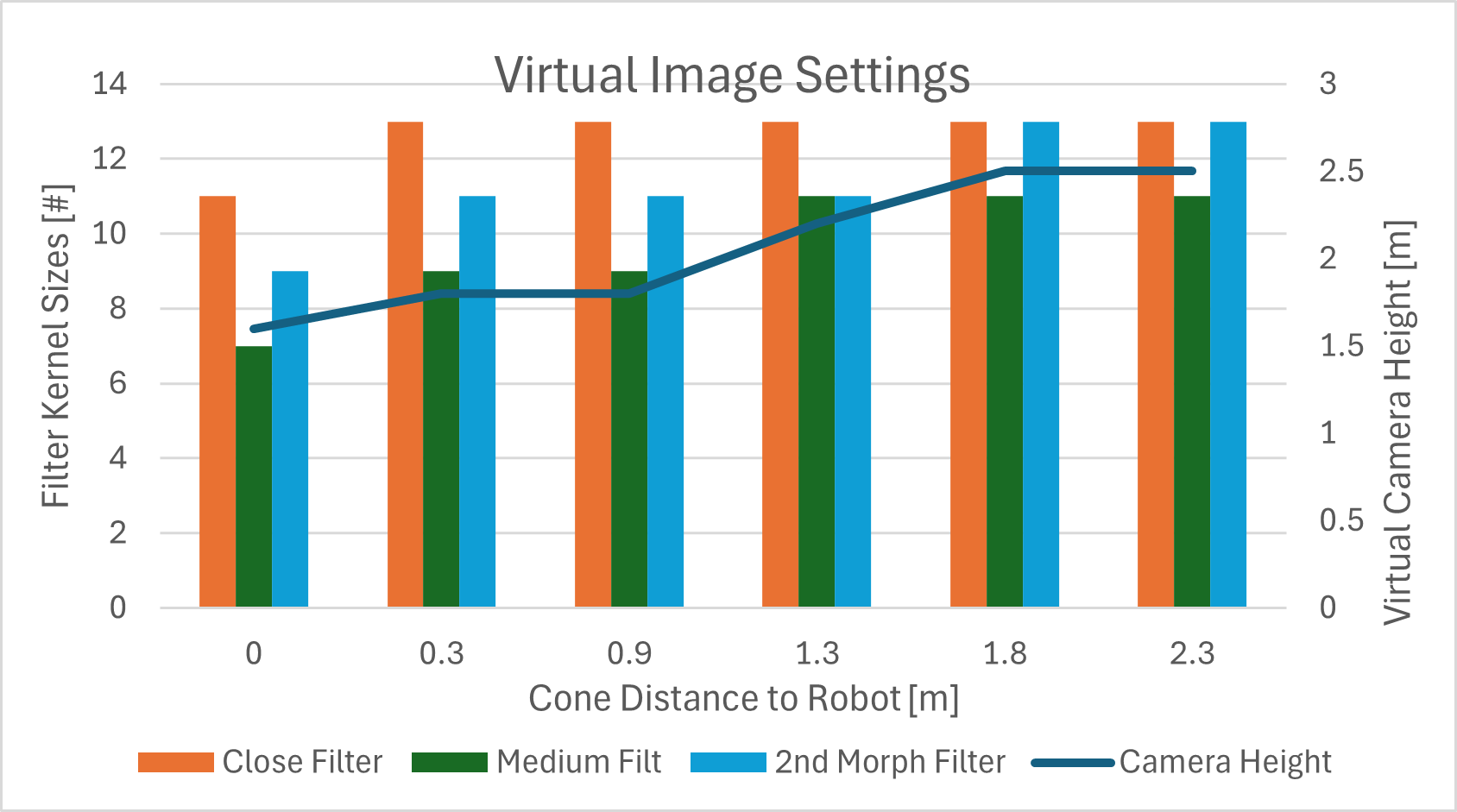}
			&
			\includegraphics[width=0.30\linewidth,height = 6 cm]{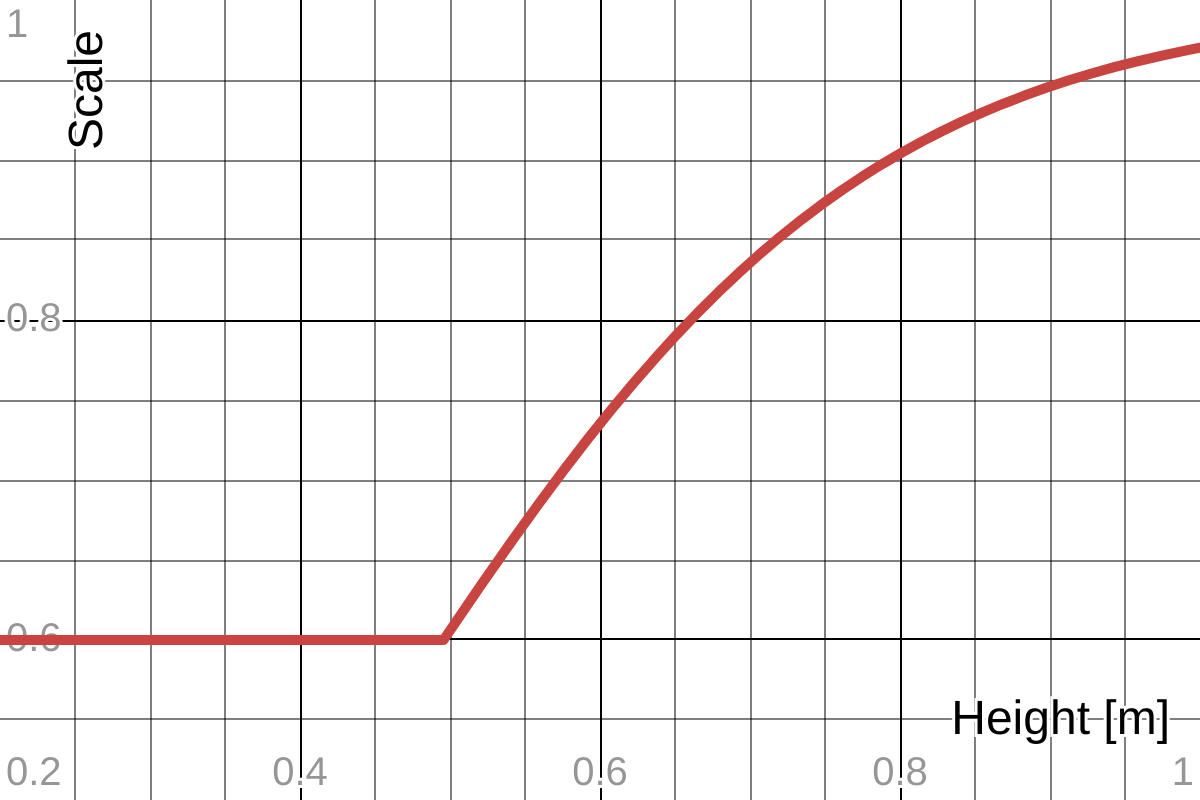}\\
			\begin{tabular}{@{}l@{}l@{}}(a) camera heights/filter size follow an increasing trend\\ 
				as the cone distance increases
			\end{tabular}
			& (b) \begin{tabular}{@{}l@{}l@{}}
				Plot of scaling Equation (\ref{eq:scale_cam_height}) that maps \\the cone height's to the camera height.
			\end{tabular} \\
		\end{tabular}
	\caption{A cone's distance and height has a shaping effect on the virtual camera settings and smoothing fiter size}
	\label{fig:dist-height-filter}
\end{figure*}

Due to beam divergence, LiDAR coverage on the cone face can vary significantly at different distances.
Further complicating the situation, the holes are not perfectly cylindrical but have a funnel-shaped opening, which narrows down to a cylinder about 30cm below the ground, which then extends to a depth of several meters. These factors can cause significant hole imprint variation in the 2D images.
To this end, we apply a three-way proximity-based camera setting adjustment, which adaptively configures virtual camera height, field of view and image processing filters to match the changing resolution of the data. We define a look-up table (Figure \ref{fig:dist-height-filter}) to model the variation of optimal camera settings as a function of cone distance.


\subsubsection{Point Sparsity}
This stems from the fact that the density of LiDAR points covering the cone varies with distance. Far away cones with larger beam spacing lead to large inter-pixel voids. Closer cones reflect a denser beam pattern, and hence reduced porosity; see Figure \ref{fig:dist-vs-camera} (a-c) for illustration.
For ease of image segmentation, a smooth cone face object is desired where all sparsity-induced voids are filled. To achieve this, we apply a number of morphological and smoothing filters (Section \ref{sec:detect-hole}), the configurations of which are defined in the lookup table in Figure \ref{fig:dist-height-filter}, where the kernel sizes follow an increasing trend as the cone distance increases.

\subsubsection{Cone face coverage}
To successfully locate the hole, it is desirable to expose the cone face as much as possible, especially at far distances. This is achieved by increasing the camera field of view such that more points are encompassed within the image frame.

\subsubsection{Hole Size}
\label{sec:track-hole-size}
The cone distance affects the extent of hole occlusion -- the hole void is more likely to be missed when the robot is distant. This, together with the hole's funnel shape can cause the hole imprint to appear larger at long distances, narrowing down towards a circle as the robot gets closer.
As shown in Figure \ref{fig:dist-vs-camera}, increasing the camera height for far away cones can mitigate this problem.

\subsubsection{Cone Height}
At the same distance, high cones can cause further hole occlusion, resulting in more pronounced variation in hole footprint; see Figure \ref{fig:dist-height-filter}(a) for an illustration. 
We apply further camera height adjustment, shown in Eq. (\ref{eq:scale_cam_height}), to ameliorate this effect,
\begin{equation}
z_{cam}^{s} =  z_{cam}  \times \textrm{max}(1 - \frac{0.9}{1+ \textrm{exp}(6.25h-2.88)}, 0.6), 
\label{eq:scale_cam_height}
\end{equation}

\noindent where parameters have been chosen to produce the profile in Figure \ref{fig:dist-height-filter}(b). This gives large camera depth to high cones, ensuring sufficient hole inclusion in the projected image.

Both the lookup table and the scaling function are easy to generate. They can be configured based on LiDAR FoV and beam number rating, and by online observation of intermediate hole detection results, as shown in Figure \ref{fig:hole-over-view}.

\section{Hole Detection}
\label{sec:detect-hole}
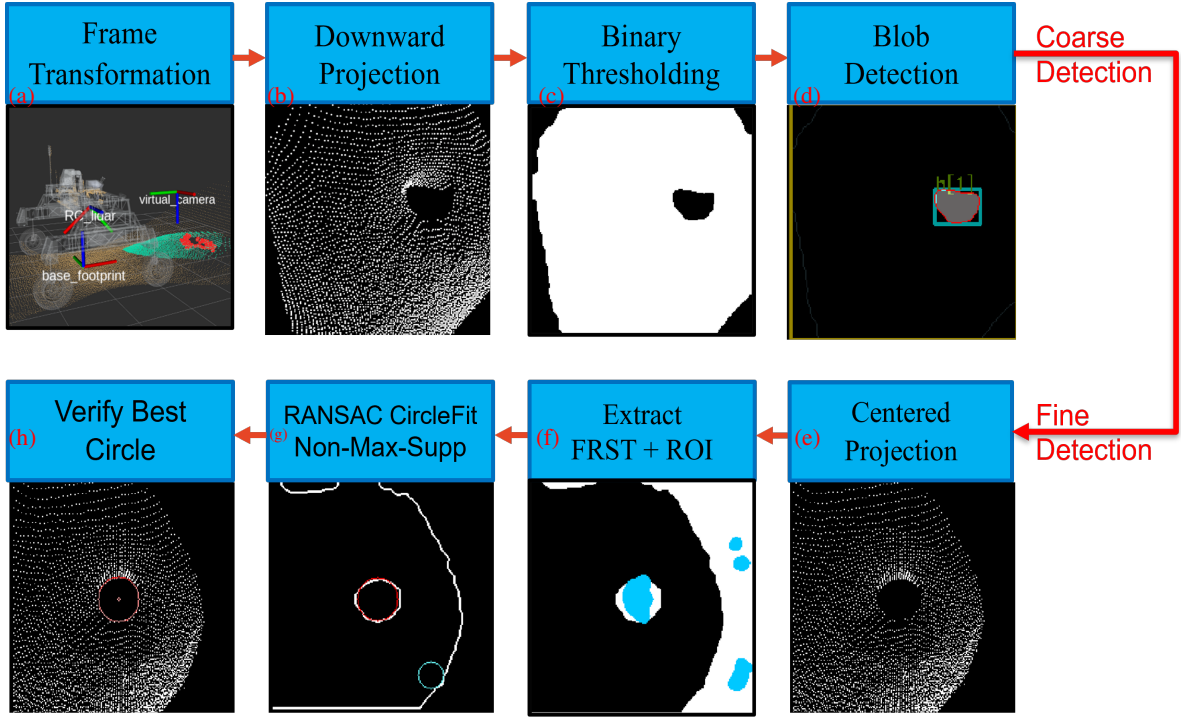
\begin{figure}[t]
	\centering		
		\begin{overpic}[width=0.95\linewidth,height=9.5cm]{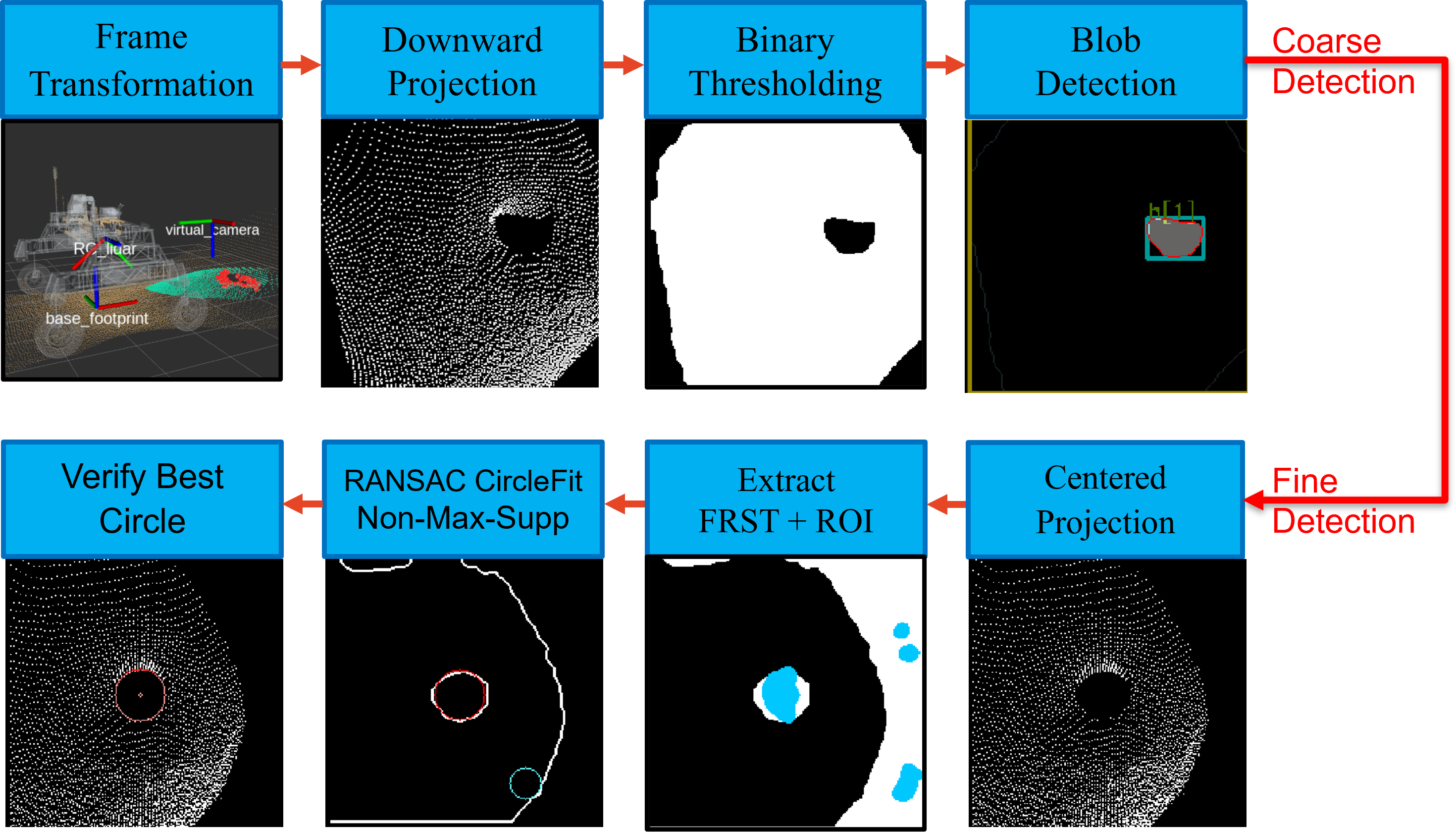}
			\put (0.3, 52) {\small{\color{red}{(a)}}}
			\put (22.2, 52) {\small{\color{red}{(b)}}}
			\put (45.2, 52) {\small{\color{red}{(c)}}}
			\put (66.8, 52) {\small{\color{red}{(d)}}}
			\put (66.9, 23.0) {\small{\color{red}{(e)}}}
			\put (45, 23) 
			{\small{\color{red}{(f)}}}
			\put (22.4, 23.5) {\tiny{\color{red}{(g)}}}
			\put (0.3, 23.0) {\small{\color{red}{(h)}}}
		\end{overpic}
	\caption{Single Shot Hole Detection}
	\label{fig:hole-over-view}
\end{figure}

Due to the data availability concerns outlined in Section \ref{sec:intro}, we elected to 
adopt a classic detection approach with hand-crafted feature extraction and probability analysis for optimal candidate selection. This approach builds on our previous work \cite{liyang2022} of two stage coarse-to-fine detection pipeline. The projection point for the virtual camera across all stages is determined by cone centre identified by the coarse stage detection. The fine detection stage then only activates when the robot is in close-proximity to the target. The entire detection pipeline is shown in Figure \ref{fig:hole-over-view}, a description of the steps is listed below

\begin{enumerate}
	\item Apply coarse projection using cone centroid as projection point, which usually does not coincide with the hole centre and gives a distorted view of the circle, as shown in Figure (b);
	\item Apply morphological operators and Gaussian blur filters to fill up spacings from sparse points. After binary thresholding, a well connected whole cone object appears and is segmented from the hole, see Figure (c)
	\item Identify the hole object with simple moment analysis, see Figure  (d)
	\item  Apply fine-stage projection using blob centre as projection point, the hole object is in the new image centre and is very circular, see Figure (e), 
	\item  Apply similar morphological operations to form connected cone object, then extract features to identify circular regions, see Figure (f). 
	\item With optimal circle fitting and Non-Maximum-Suppression, we identify the best circle in Figure (g). 	
\end{enumerate}

\subsection{Coarse Detect}
\label{sec:coarse-detect}
In this stage, standard contour methods are applied for hole detection. 
Since the cone and hole objects have fairly consistent sizes, simple thresholding can yield good detection results. Of all extracted contours, the largest one is usually the cone, shown in Figure \ref{fig:hole-over-view} (d) as a golden bounding box. If a void is found inside the detected cone, it is classified as the hole object if it is both of sufficient size and close to the image centre. A rough position of hole can thus be obtained after 2D-3D back projection.

\subsection{Fine Detection}
To locate precisely the position and dimension of the hole, we project the cone point cloud one more time, with the virtual camera right above the detected blob centre. This ensure the hole to appear highly circular in this second virtual image, avoiding any perspective effect.

We now present our best hole selection scheme, which is an extension of the optimal circle fitting algorithm in \cite{liyang2022}, with the addition of a non-maximum-suppression method. 


\subsubsection{FRST feature and ROI extraction}
\begin{figure}[t]
		\centering
		\begin{tabular}{@{}c@{}}		
			\includegraphics[width=0.9\linewidth]{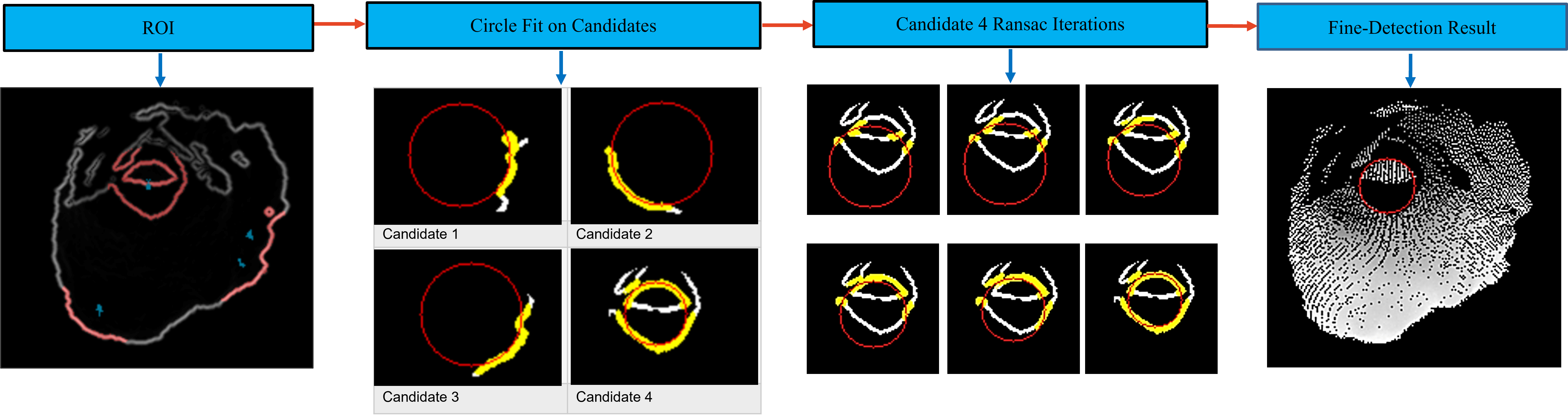}\\
			\begin{tabular}{|l|l|l|l|}
				\hline
				\begin{tabular}{@{}l@{}}		
					\small{(a)	FRST features are }  \\
					\small{the small \textcolor{cyan}{blue} blobs. }\\ \small{ROI's  pixels are \textcolor{red}{red}.}
				\end{tabular}
				& 
				\begin{tabular}{@{}l@{}}
					\small{(b) RANSAC on four candidates }\\
					\small{ and  their Least Squares circle} \\
					\small{ fitting score. Candidate  4  has   }\\
					\small{ the highest fitted pixel count.} \\
				\end{tabular} 
				& 
				\begin{tabular}{@{}l@{}}
					\small{(c) A series of RANSAC itera-} \\
					\small{tions  in the 4th candidate.}\\
				\end{tabular} 
				&
				\begin{tabular}{@{}l@{}}
					\small{(d) Fine-detection result:} \\
					\small{overlay the best circle on} \\
					\small{ original depth image.}\\
				\end{tabular} 
				\\
				\hline
			\end{tabular}\\
		\end{tabular}
	\caption{FRST + ROI + RANSAC + Taubin}
	\label{fig:frst+ransac}
\end{figure}

A Sobel filter is applied to the output of the morphological operation and a gradient image is produced, showing clear edges of the virtual image. 	We use the hand-crafted Fast Radial Symmetrical Transform (FRST) \cite{FRST} features to detect circular regions. The FRST feature behaves like a 2D histogram, high counts indicate the possible locations of circle centre. The user is advised to refer to Appendix \ref{app:FRST-extraction} for further details. Pixels within an acceptable neighbourhood of the feature centre are referred to as Regions of Interest, circles can be identified from the extracted ROI's.
Figure \ref{fig:frst+ransac}.(a) gives an illustration of this process, FRST feature points are in blue blobs, and ROI's extracted are in red pixels. For the sake of completeness, we present the whole procedure.

\subsubsection{RANSAC and Circle Fitting}
\label{sec:ransac-circle-fit}
Given a set of points that form a circle, we opt to use Taubin's Least Squares Fitting algorithm \cite{taubin-error-analysis} to find the optimal circle. For the sake of completeness, we list the problem statement here. 

Let us denote $\mathbb{R}$ as the set of ROI's in given image $I$.  Then denote $R$ as the set of valid  pixels  $R=\{\mathbf{p}\}$ with size $\lvert R \rvert$ from the the FRST generated ROI. And let $\bm{\Theta} = (\mathbf{c}, \; r)$ denote the circle fitted to the ROI with centre $\mathbf{c}= (a, b)$ and radius $r_i$, the ROI points  $\{x_i,\, y_i\}$ spreat around the circle are described as \\
\begin{equation}
\begin{aligned}
\cos \phi_i &= \frac{x_i + \delta_i - a}{r} = \frac{x_i - a}{r} + \tilde \delta_i, \quad \tilde \delta_i = \frac{\delta_i}{r} \in  \mathcal{N}(0, \sigma)\\ 
\sin \phi_i &= \frac{ y_i + \epsilon_i - b  }{r} = \frac{y_i - b}{r} + \tilde \epsilon_i, \quad \tilde \epsilon_i = \frac{\epsilon_i}{r} \in  \mathcal{N}(0, \sigma). \\
\end{aligned}
\end{equation}

\begin{equation}
\begin{aligned}
\text{Now solve } & \bm{\Theta} = \underset{}{\text{argmin}} \, \mathcal{F}_{\mathrm{T}} (\bm{\Theta}) \,\\
\text{where } \mathcal{F}_{\mathrm{T}}	&=  
\frac{\Sigma[ (x_i - a)^2 + (y_i - b)^2 - r^2 ]^2}
{4 n^{-1} \Sigma[ (x_i - a)^2 + (y_i - b)^2 ]}, \quad n = \lvert \mathbb{R} \rvert.\\
\end{aligned}
\end{equation}

A change of parameters transforms above into a linear least squares problem
\begin{equation}
\begin{aligned}
& \mathcal{F}_{\mathrm{T}}	=  \Sigma
\frac{[ A z_i + B x_i + C y_i +D ]^2}
{ n^{-1} \Sigma[ 4 A^2 z_i + 4A B x_i + 4A C y_i + 4B^2 + 4C^2  ]},\\
\text{where } & z_i = x_i^2 + y_i^2, \quad B = - 2aA, \quad
C = -2bA, \quad
D = a^2 + b^2 - r^2\\
\end{aligned}
\end{equation}

\begin{equation}
\begin{aligned}
\text{This } & \text{is equivalent to the following objective function} \\
& \mathcal{F}_l = \sum_{i=1}^n (Az_i + Bx_i + Cy_i + D)^2, \\
\text{subject to: } & 4A^2 M_z + 4ABM_x + 4ACM_y + B^2n + C^2n = 1, \\
\text{where }& M_x =\sum_{i=1}^n x_i. \\
\label{eqn:Taubin}
\end{aligned}
\end{equation}

This circle fitting problem can now be transformed into a Eigen-decomposition based approach which is Linear and has an analytical solution. Details are given in Appendix \ref{app:taubin}. Taubin's algorithm has an important property \cite{taubin-error-analysis}, the solutions are independent of the choice of the coordinate frame, i.e. invariant under translations and rotations. it fully respects the circular hole's radial symmetry,  without coercing for a meaningless orientation. The residuals are Identically and Independently Distributed (IID) around the smooth circumference of the circle. This contrasts with other non-ideal methods such as in \cite{bhole-svm}  where a circle is treated as a polygon  and piece-wise linear HOG features are  extracted around the hole circumference then coerced to describe a circular geometry via non-linear SVM regression. Such methods would require large amount of data augmentation in order to cater for all orientations subject to finite angular resolution. 


Since not all points in the ROI are part of the circle, we separate the valid pixels and outliers in a RANSAC \cite{RANSAC} framework based on Taubin. Starting with a seed of trial samples, the process first finds the optimal fitted circle, then grows this circle by adding more points that satisfy the current constraint. This process repeats until circle growth converges or max iteration number is reached. Since Taubin algorithm is a type of Convex optimsation, each iteration is guaranteed to converge to the optimal fit. This ensures RANSAC process leads to the largest optimal circle in an ROI.
The result of RANSAC for each ROI is illustrated in Figure \ref{fig:frst+ransac}(b)-(d). Since RANSAC is a very expensive operation, we set a cap of 200 on the number of retries in the candidate RANSAC process, to balance between fitting accuracy and visual servoing delay. 

Our method provides highly accurate detection results, yet is fast to compute in real-time, without incurring expensive data learning based process or expansive hardware. It is therefore highly suitable for an in-situ robot inspection application.

\subsubsection{Non-Maximum-Suppression}
\label{sec:n-m-s}
The previous steps can produce multiple candidate circles, see Figures \ref{fig:hole-teaser}(f), \ref{fig:frst+ransac}(b) and \ref{fig:hole-over-view}(g). This is exacerbated when phantom holes appear at non-vertical LiDAR scan as Convex Hull cannot totally eliminate them. We provide a Non-Maximum Suppression (NMS) scheme to select the best candidate. This scheme takes into consideration the following factors: FRST feature strength, the candidate's distance to image centre and circular goodness of fit. 

\noindent Before presenting any rigorous analysis, we first define some simple switching logic to  discard invalid candidates. This is analogous to the Relu operation in a NeuralNet system:
\begin{itemize}
	\item Circle radius is in a prescribed range 
	\item Circularity score (Eq. \ref{eq:score_circle}) is above a threshold
	\item More than 70\% of region is black
	\item Centrality score (Eq. \ref{eq:score_reg}) above threshold
	\item FRST feature score (Eq. \ref{eq:score_frst}) above threshold
\end{itemize}

Continuing from the problem statement in Section \ref{sec:ransac-circle-fit}. We re-use the notations here: denote $R_i$ as the set of valid  pixels  $R_i=\{\mathbf{p}_i\}$ with size $\lvert R_i \rvert$ from the the $i$'th FRST generated ROI. And let $\bm{\Theta}_i = (\mathbf{c}_i, \; r_i)$ denote the circle fitted to the ROI with centre $\mathbf{c}_i$ and radius $r_i$. To find the best circle $\Theta^*$, we first define the occurrence probability $P(\Theta_i \vert I)$ of the $i$'s candidate circle. 
\begin{equation*}
\begin{aligned}
P(\bm{\Theta}_i \vert I) &= \Sigma_r P(R_r \vert I)  P(\bm{\Theta}_i \vert R_r)  \quad\quad\quad\quad\, \leftarrow\text{\small{Marginalize}}\\
&= P(R_i \vert I)  P(\bm{\Theta}_i \vert R_i)  \quad\quad\quad\quad\quad\;\; \leftarrow\text{\small{Simplify}}\\
&\propto P(R_i \vert I)  P(\bm{\Theta}_i)  P( R_i \vert \bm{\Theta}_i)  \quad\quad\;\;\;  \leftarrow\text{\small{Bayes rule}}\\
&\propto P(R_i \vert I) P(\bm{\Theta}_i)  
\prod_{\mathbf{p}_j \in R_i} P( \mathbf{p}_j \vert \bm{\Theta}_i )  \;  \leftarrow\text{\small{Likelihood}}
\end{aligned}
\end{equation*}

This best circle $\bm{\bm{\Theta}^*}$ should correspond to the highest probability of occurrence, or maximum a posteriori estimate. This notion allows us to define the  confidence score $\mathcal{S}_{conf}(\bm{\Theta}_i)$ on candidate $\bm{\Theta}_i$ as a weighted sum of the ROI quality score $\mathcal{S}_{F}(R_i)$, regularization score $\mathcal{S}_{reg}(\bm{\Theta}_i)$ and circularity fit score $\mathcal{S}_{circle}(\bm{\Theta}_i)$.
\begin{equation}
\mathcal{S}_{conf}(\bm{\Theta}_i) = \alpha_1 \mathcal{S}_{F}(R_i) \,
+ \, \alpha_2 \mathcal{S}_{reg}(\bm{\Theta}_i) \,
+ \, \mathcal{S}_{circle}(\bm{\Theta}_i).
\end{equation}
\noindent The best circle is the one with highest confidence:
\begin{equation}
\bm{\Theta}^* = \underset{i \in \lvert \mathbb{R} \rvert}{\text{argmax}} \,  \mathcal{S}_{conf}(\bm{\Theta}_i)
\end{equation}

\noindent The ROI quality is determined by the FRST feature quality, high feature count $\lvert F_i \rvert$ strongly implies existence of circle: 
\begin{equation}
\mathcal{S}_{F}(R_i) = \frac{1}{1 + 3 \exp(3-0.1\lvert F_i \rvert)} \quad \in [0 \sim 1] 
\label{eq:score_frst}
\end{equation}

\noindent Despite the convex hull filtering in Section \ref{sec:cone-detect},
phantom holes can still appear in the cone face due to partial LiDAR scan occlusion at oblique angles.
Using the fact that bore holes are located at the blast cone centre, we add a regularisation term to penalise phantom holes that are further from the apparent cone centre,
\begin{equation}
\mathcal{S}_{reg}(\bm{\Theta}_i) = \frac{1.2}{1+50\exp(0.05 \norm{ \mathbf{d}_i}-5.5)} \quad \in [0 \sim 1],
\label{eq:score_reg}
\end{equation}
where $ \mathbf{d}_i =  \mathbf{c}_i-\mathbf{o}_I $ is the distance vector between circle centre to image centre $\mathbf{o}_I$.

To evaluate how well the ROI points fit a circular hole, it is necessary to consider both residual errors and the ROI's angular coverage on the full circle. 
We define a circular grid of $B$ bins around the hole to sort points into its pertaining bins from their angles of incidence. We compute each bin's mean-squared-error $e_b$, then map it to a bin probability score $H_b$. The circularity score is the sum of individual bin scores:
\begin{alignat}{2}
\mathcal{S}_{circle}(\bm{\Theta}_i)  & =\; \frac{1}{B}\sum_{b=1}^B H_b \quad \in [0 \sim 1],
\label{eq:score_circle} \\
\intertext{where, bin score $H_b$ is a type of probability mapped from bin Mean-Squared Error (MSE): }
H_b =\; &\text{exp}(-\frac{1}{2}\frac{e_b}{\sigma^2} ) \quad \in [0 \sim 1], \\
\intertext{and, bin MSE $e_b$ accumulates all point errors in bin $b$ :} 
\forall \, j \in \lvert R_i \rvert, \; &  	\mathbf{v}_j =  \frac{1}{r_i}(\mathbf{p}_j - \mathbf{c}_i) ,    \nonumber  \\
& e_j = \norm{ 1 - \norm{\mathbf{v}_j} }^2, \quad
b = \text{Bin}( arctan( \mathbf{v}_j ) ),   \nonumber  \\
&{e_b}^+ = \frac{{n}_b^{-} {e}_b^{-} +   e_j }{\hat{n}_b^{-} + 1}, \quad
n_b^{+} = n_b^{-} + 1.
\label{eq:hist-bayes}
\end{alignat}\\


\section{Experiments}
\label{sec:experiment}


We test DIPPeR’s navigation and perception systems with both simulation and real-world data. For simulation, we build simulation world in Gazebo with 3D terrain mesh from various sources. Similar to our previous work \cite{liyang2020}, We added hollow cylinders with prescribed diameters in the 3D mesh to resemble blast holes. For real-world tests, we performed extensive tests on the DIPPeR platform. This includes numerous functional tests on Sydney University Campus and two site-trials at blast-sites of mines in Australia and USA.

\subsection{DIPPeR Design}
\subsubsection{Hardware Configuration}
The DIPPeR robot is a 3.8m-wide, 2.4m-long omnidirectional platform with 1.3m of ground clearance, capable of in-place rotation. Weighing approximately 320 kg, it achieves speeds of up to 1 m/s. 

The platform is equipped with two LiDAR sensors: a 32-beam Ouster OS-1~\cite{ouster1-32} for situational awareness and cone detection, and a 128-beam Ouster OS-0~\cite{ouster0-128} for high-precision hole detection. 
The DIPPeR platform uses an Advanced Navigation Certus dual antenna GNSS/INS~\cite{certus-ins}, capable of achieving 
+/- 1.2m accuracy in absence of RTK. Certus also includes a 9-DOF IMU device, which  uses high-precision, temperature-stabilised MEMS sensors measuring linear acceleration, angular velocities, and magnetic field strength. These are fused with the GNSS solution to provide roll, pitch, and heading accuracy of 0.1 degrees. It also serves as the GNSS-disciplined clock for our PTP and NTP network time synchronisation.
The On-board computing platform is a 12-core Intel NUC 13 computer, it hosts all of DIPPeR's software modules that work seamlessly with the hardware enabling DIPPeR to run inspection continuously in realtime.

\subsubsection{Software Architecture}
DIPPeR's software modules include the \ac{GNC} system, and the perception system. Most of DIPPeR's software was written in C++ and Linux scripts, using ROS Noetic \cite{ROS-09}, running on Ubuntu 20.04. 

The \ac{GNC} system makes use of the flexible and extensible \texttt{move\_base\_flex}\footnote{\url{http://wiki.ros.org/move\_base\_flex}} package, which, in addition to backwards compatibility with the standard \ac{ROS} \texttt{move\_base}, also exposes a number of useful actions that allow for more complex interactions with motion controllers and planners. We specifically use A$^*$ for the global path planning and the \ac{TEB} planner \citep{Rosmann2017} for local trajectory planning. We additionally developed a \ac{ROS} controller for omnidirectional swerve-steer platforms such as DIPPeR, which executes twist commands from the navigation system while satisfying steering angle constraints.

The global and local path planners are invoked during the global and fine planning stages of the mission, operating in the UTM and Odom frames respectively. The robot's UTM coordinates are accessible directly from  the Certus INS. We obtain an estimate of the robot local odometry by running an EKF ~\cite{robot-Ekf-2014} in local mode to fuse the wheel odometry and the IMU. Upon transition to the fine positioning stage, visual servoing on the detected hole centre is performed, transforming the position from the virtual camera frame to the robot body frame to determine the positional offset, and then driving the robot's sonde axis towards that location. This process fully exploits the pseudo-omnidirectionality of the platforms motion to position the body of the robot precisely for the subsequent down-hole sensing.

We developed our proprietary decision-making and target detection system in C++. For 3D LiDAR processing we make extensive use of the PCL library~\cite{PCL-2011}, and in 2D hole detection, we use both the OpenCV library~\cite{opencv-library} for basic image processing and the circle\_fit library~\cite{git-circle-fit} for implementation of Taubin's algorithm. 
The perception system is capable of processing three hole detections per second, which at the operating speeds of the platform is sufficient to perform real-time hole inspection.

\subsection{Unit Test}
\subsubsection{Robustness Analysis -- on Real-world Data}
\begin{figure}[h]
	\centering
	\setlength\tabcolsep{2pt}	
	\begin{tabular}{@{}cc@{}}
		\begin{overpic}[width=0.37\linewidth]{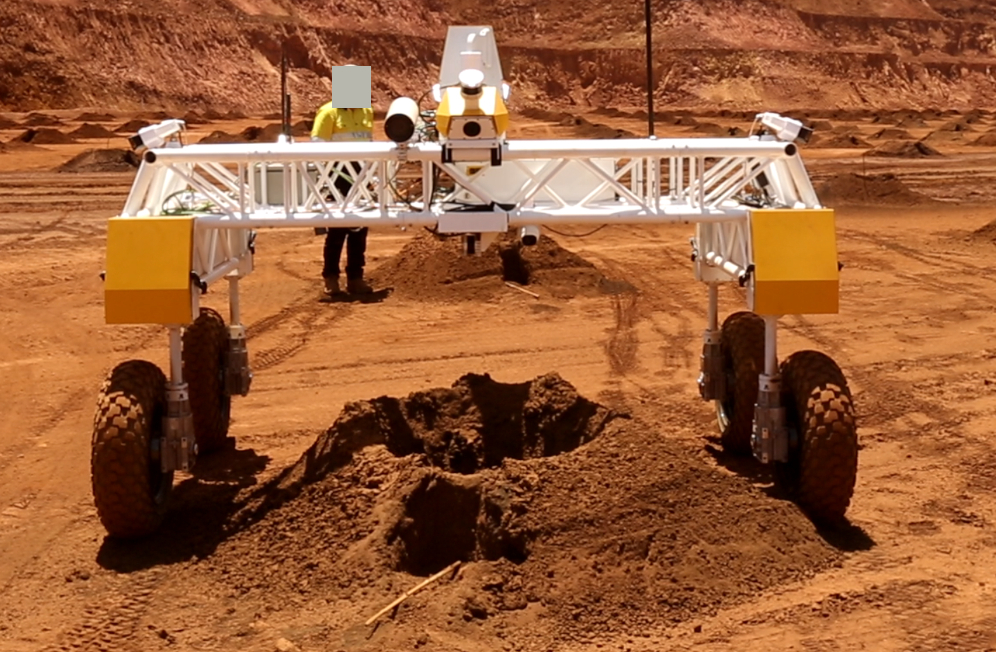}
			\put (0.1, 58)
			{\small{\color{red}{(a)}}}	
		\end{overpic}   
		&
		\includegraphics[width=0.55\linewidth]{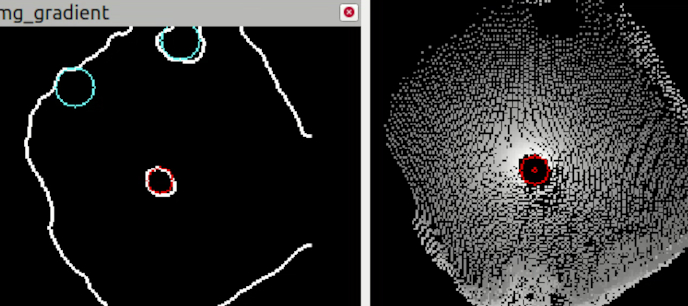}
		\\	
		\begin{overpic}[width=0.37\linewidth]{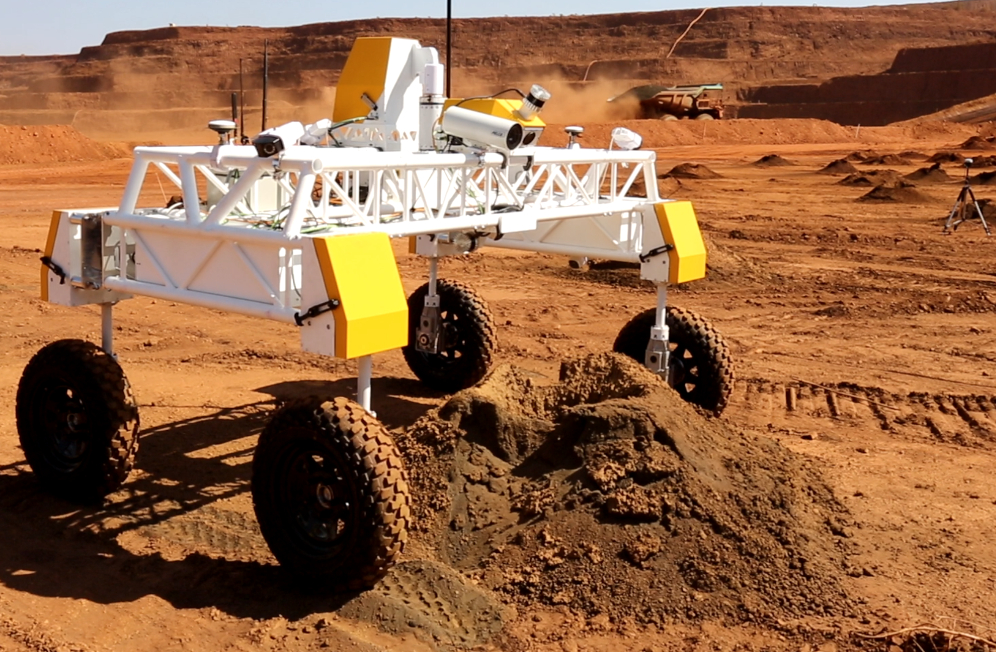}
			\put (0.1, 58)
			{\small{\color{red}{(b)}}}	
		\end{overpic}   
		&
		\includegraphics[width=0.55\linewidth]{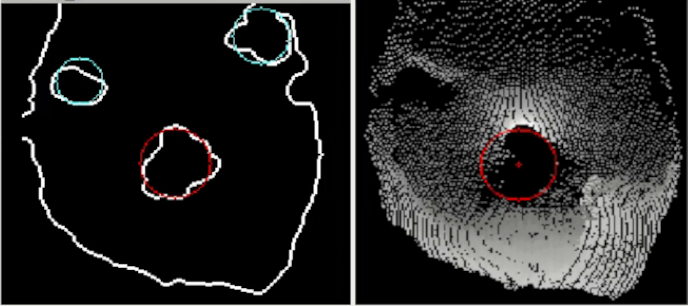}
		\\		
		\begin{overpic}[width=0.37\linewidth]{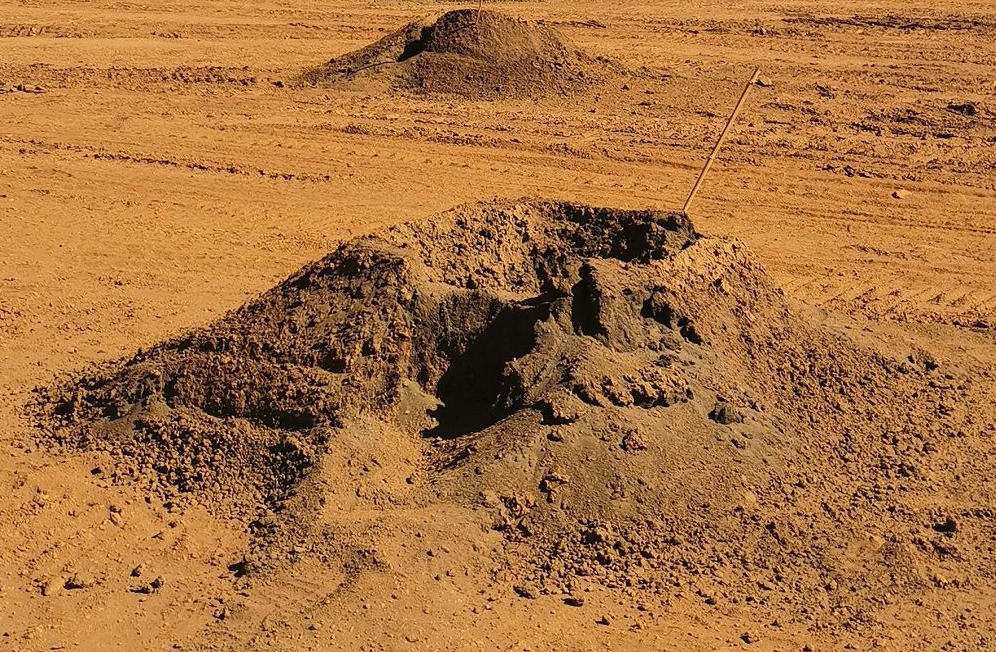}
			\put (0.1, 58)			{\small{\color{red}{(c)}}}	
		\end{overpic}   			
		&
		\includegraphics[width=0.55\linewidth]{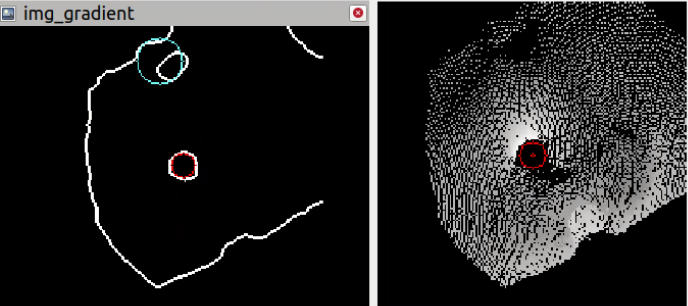}
		\\
		\begin{overpic}[width=0.37\linewidth]{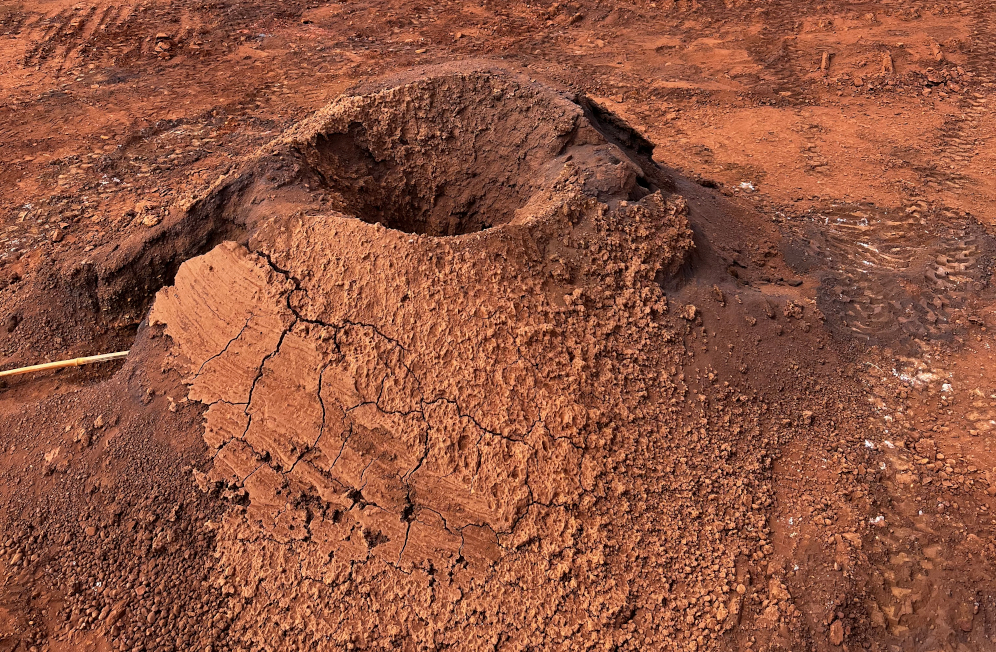}
			\put (0.1, 58)			{\small{\color{red}{(d)}}}	
		\end{overpic}   
		&		\includegraphics[width=0.55\linewidth]{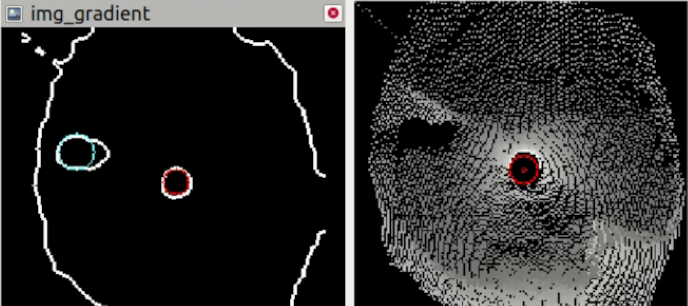}\\		
	\end{tabular}
	\caption{NMS on phantom holes. \\
		Note for left column:  \textbf{\textcopyright Rio Tinto 2025, All Right Reserved.}}
	\label{fig:notch}
\end{figure}

Combined use of Convex-Hull in 3D cone detection (Section \ref{sec:cone-detect}) and Centrality regularisation in Hole Non-Maximum-Suppression (Secton \ref{sec:n-m-s}) are effective in ruling out phantom holes. As shown in Figure \ref{fig:notch}.

\subsubsection{Cone Detection Result -- on Simulation Data}
\begin{figure}[h]
	\centering
	\begin{tabular}{@{}c@{}}
		\includegraphics[width=0.40\linewidth, height=4cm]{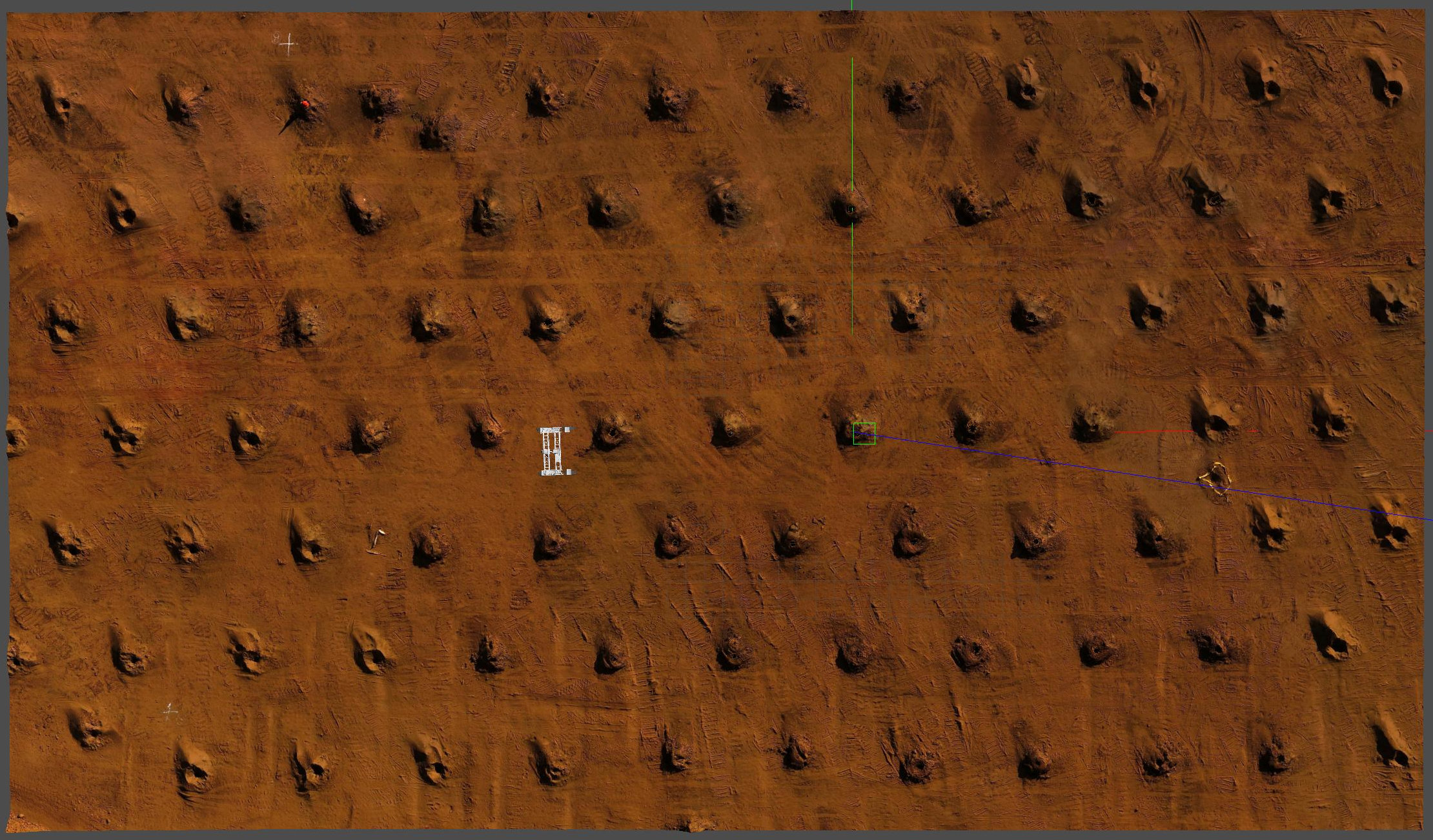} \\
		(a) A simulated world, \textbf{\textcopyright Rio Tinto} \\
		\includegraphics[width=0.40\linewidth,height=4cm]{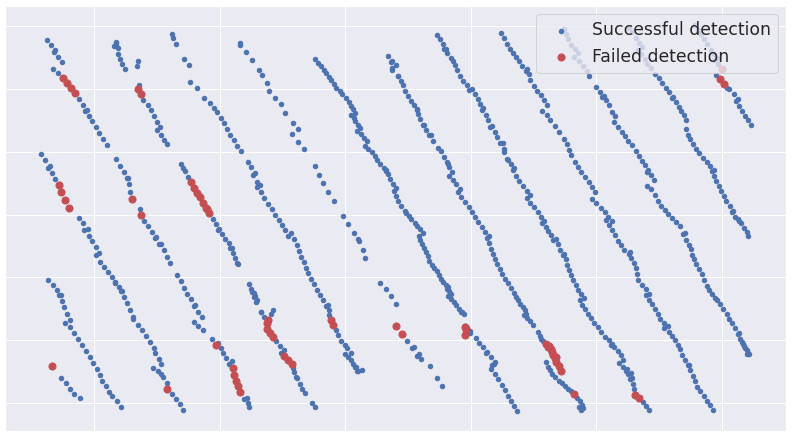}\\
		(b) Cone detection results
	\end{tabular}	
	\begin{tabular}{@{}c@{}}
		\includegraphics[width=0.5\linewidth,height=9cm]{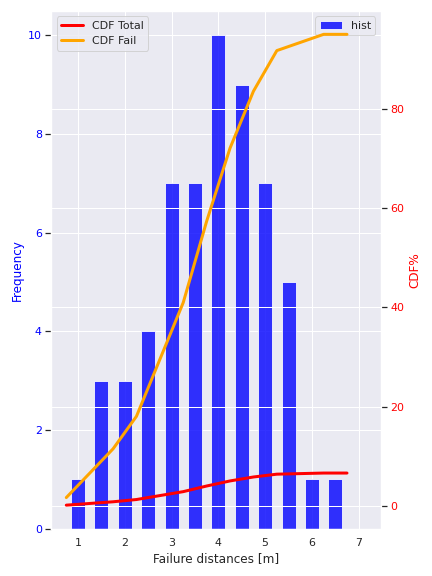}\\
		(c) Failure Histogram
	\end{tabular}
	\caption{Statistical analysis on safe cone detection distance in simulated world. \\Note for Subfigure (a): \textbf{\textcopyright Rio Tinto 2025, All Rights Reserved. 3D surface model generated using Bentley Systems, Inc. software. Visualised using 	Gazebo $[$11.0/Open-Robotics$]$.} }
	\label{fig:cone-test-results}
\end{figure}

We conducted cone detection accuracy test for the high-density LiDAR at various positions, on a simulated world using 3D mesh data generated by Bentley\texttrademark  software\cite{bentley}. About 5\% of the total detections are failure cases. The results are presented in Figure \ref{fig:cone-test-results}(a), where blue dots indicated locations of successful detection, red dots failures. We also grouped failure detection counts into histogram bins based on their distances to the robot. The failure distance histogram and converted Cumulative Distribution Function are shown in  Figure \ref{fig:cone-test-results}(b). 
From the plot high cone detection failure is observed to occur at distance range $[3\mathrm{m} \sim 6\mathrm{m}]$, whereas failure rate becomes very low once the robot distance reduces to three meters. This allows us to set activation distance for high-density lidar processing as described in Section \ref{sec:dist-range} to start at 3 meters distance.

\subsubsection{Approach Behavior Analysis -- on Real-world Data}
\begin{figure}[h]
	\centering
		\begin{tabular}{cc}
			\includegraphics[width=0.4\linewidth]{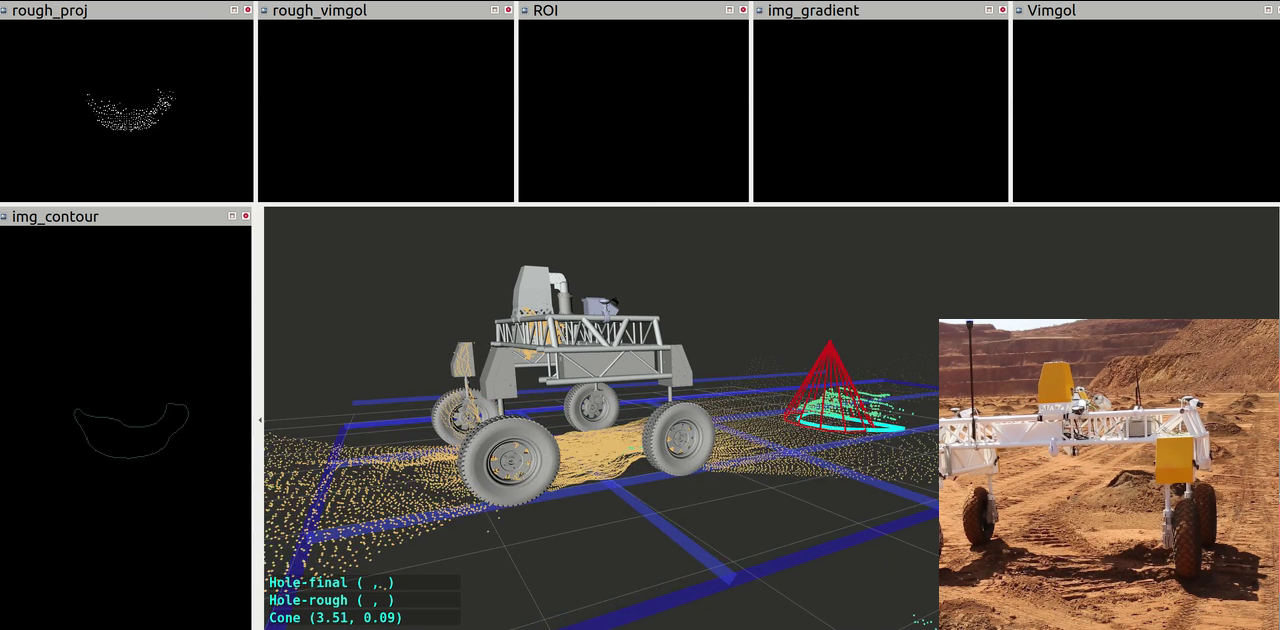} &
			\includegraphics[width=0.4\linewidth]{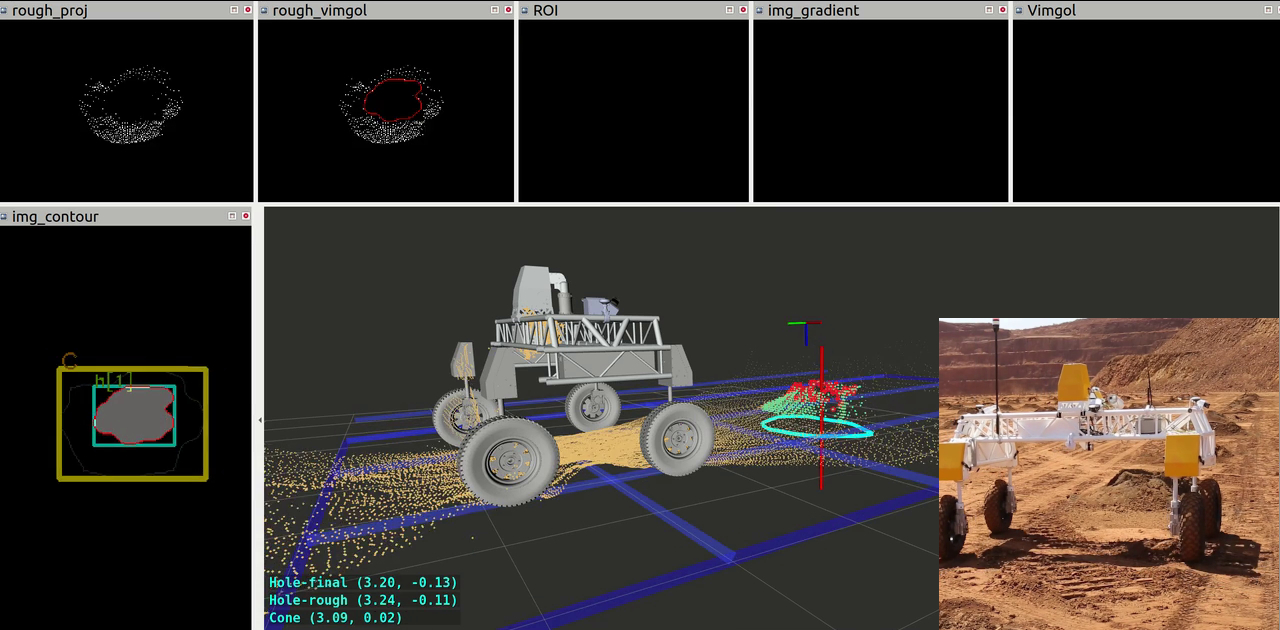} \\
			(a) Distance = 3.51 m, with sparse LiDAR &
			(b) Distance = 3.09 m \\
			\includegraphics[width=0.4\linewidth]{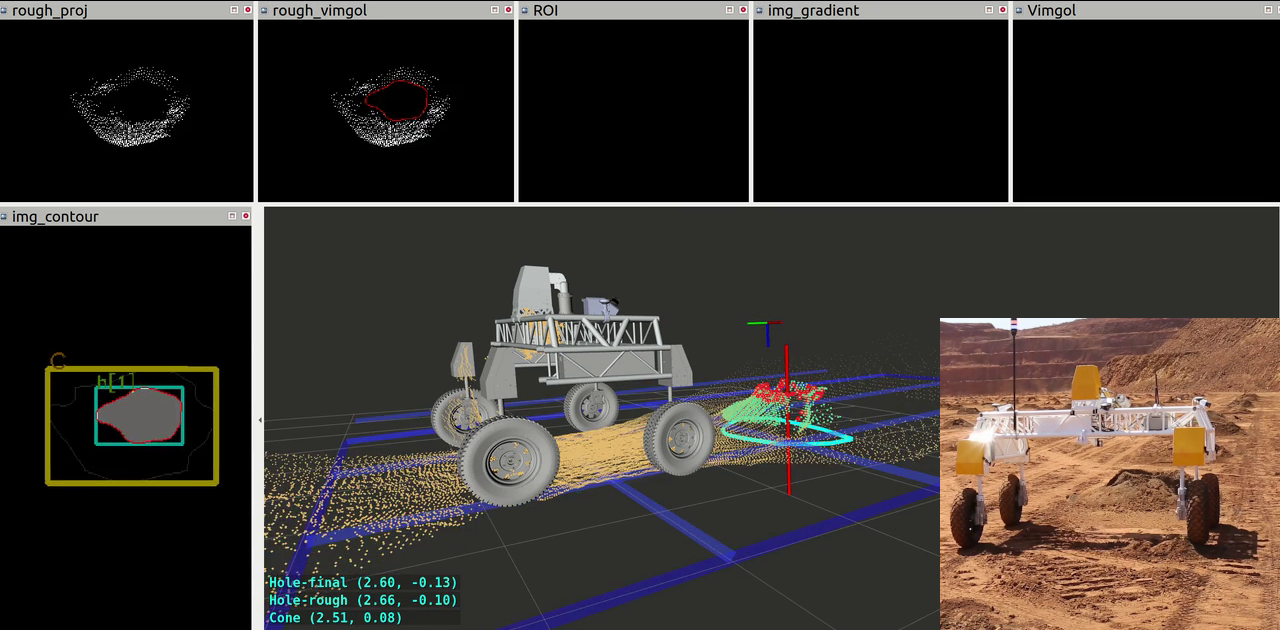} &
			\includegraphics[width=0.4\linewidth]{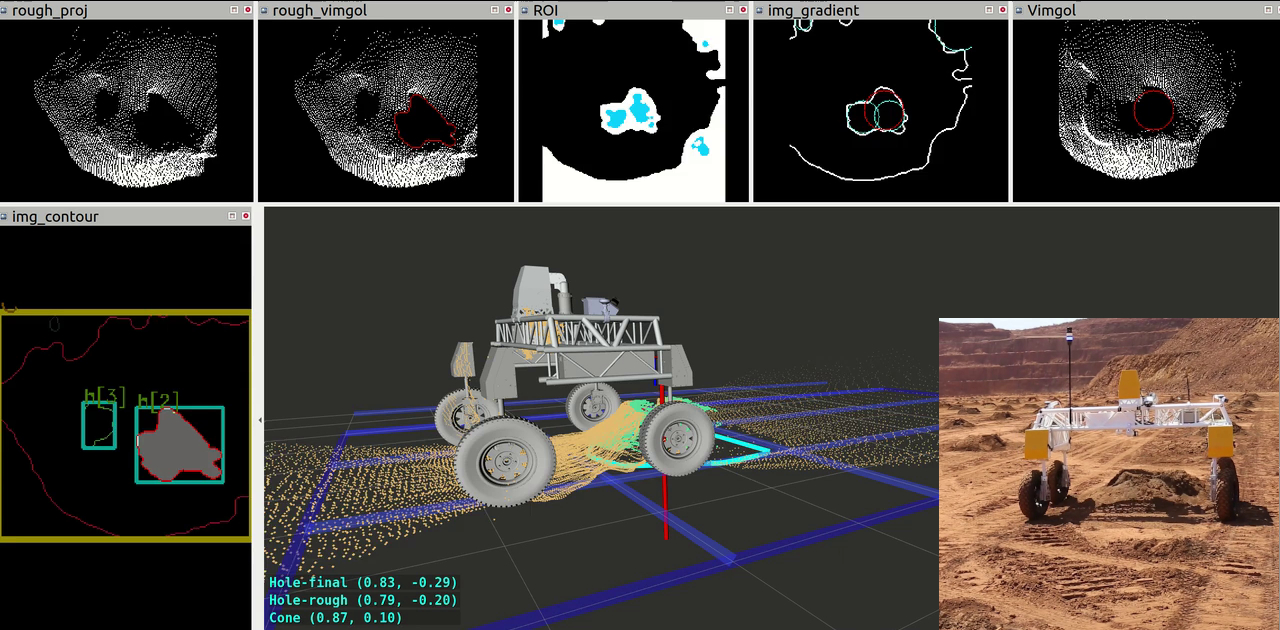} \\
			(c) Distance = 2.51 m &
			(d) Distance = 0.87 m \\
			\includegraphics[width=0.4\linewidth]{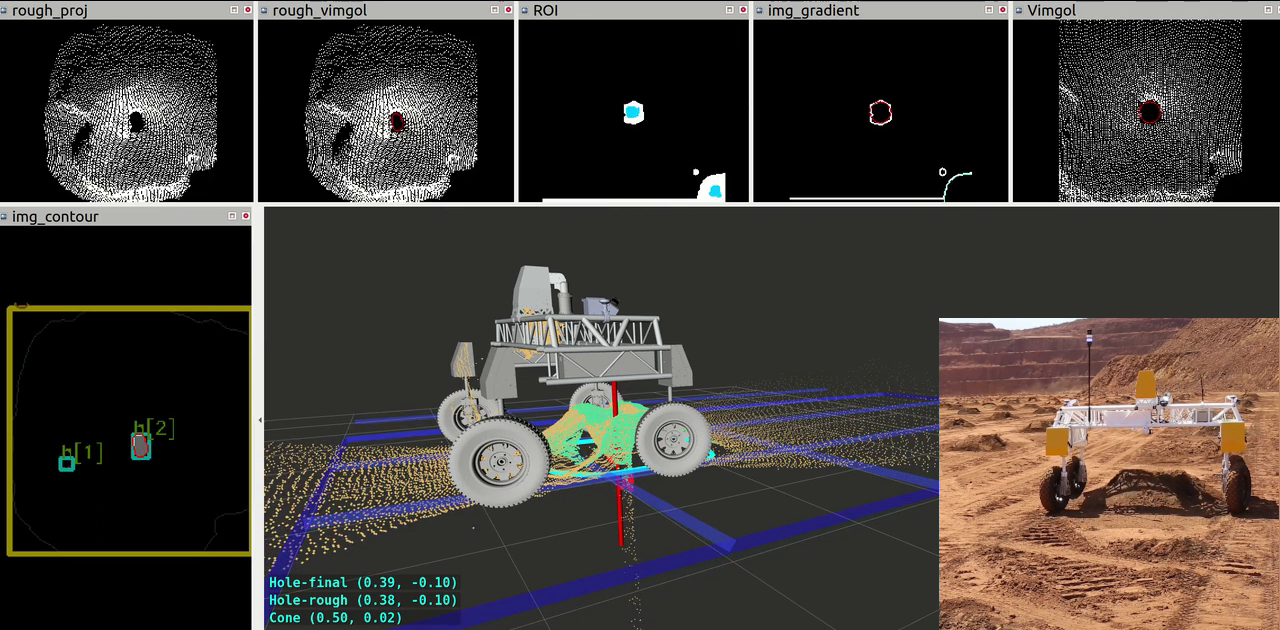} &
			\includegraphics[width=0.4\linewidth]{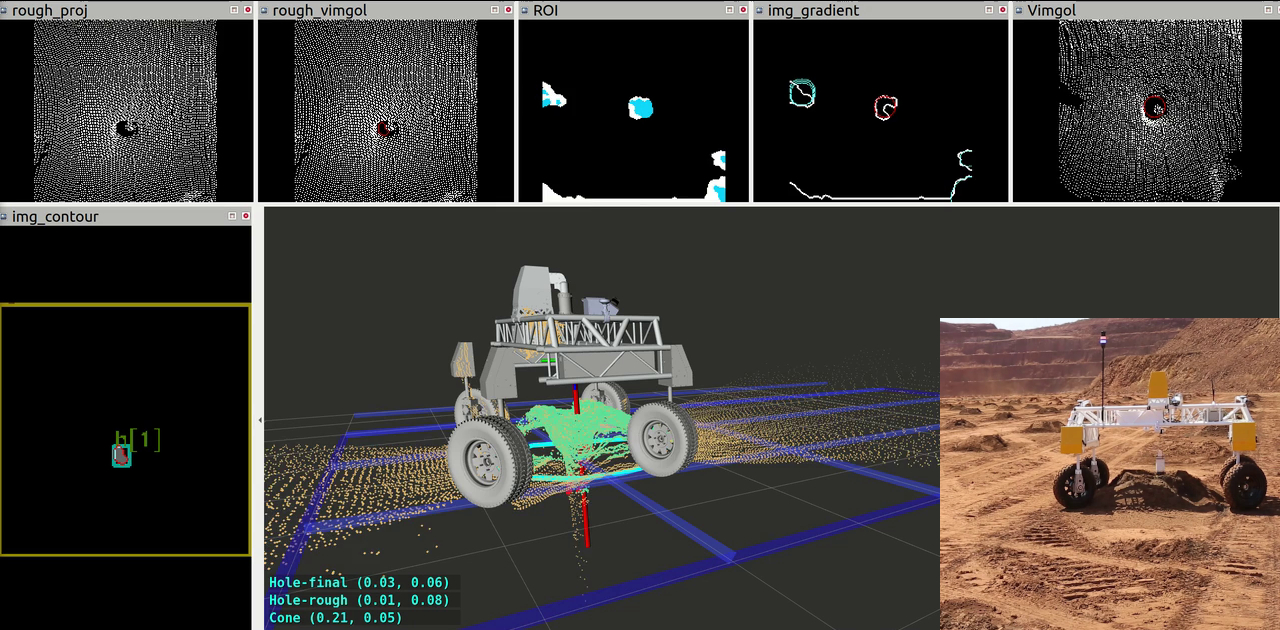} \\
			(e) Distance = 0.50 m &
			(f) Distance = 0.01 m, sonde dropped \\
		\end{tabular}	
	\caption{Approaching behaviour at various target distances. \\
		Note for all inset photos: \textbf{\textcopyright Rio Tinto 2025, All Right Reserved.}}
	\label{fig:simu-approach}
\end{figure}

We also tested our auto-adjust perception feature in physical-world test. We illustrated in Figure \ref{fig:simu-approach} below the perception results in a DIPPeR's motion sequence as it gradually approaches the cone/hole target. The detection results are shown for 6 different distances: 3.51m, 3.09m, 2.51m, 0.87m, 0.50m and 0.01m. At 3.51m distance, the robot sees a cone target from its sparse LiDAR and has just transitioned from seek-GPS-point mode. At 0.01 target distance, the robot has stopped visual servoing and is in the process of hole-dipping. The changes in cone face density and hole sizes match the expected behaviour. 

\subsubsection{Mission Test - at Usyd Campus }
\begin{figure}[h]
	\centering
	\begin{tabular}{cc}
		\includegraphics[width=0.4\linewidth]{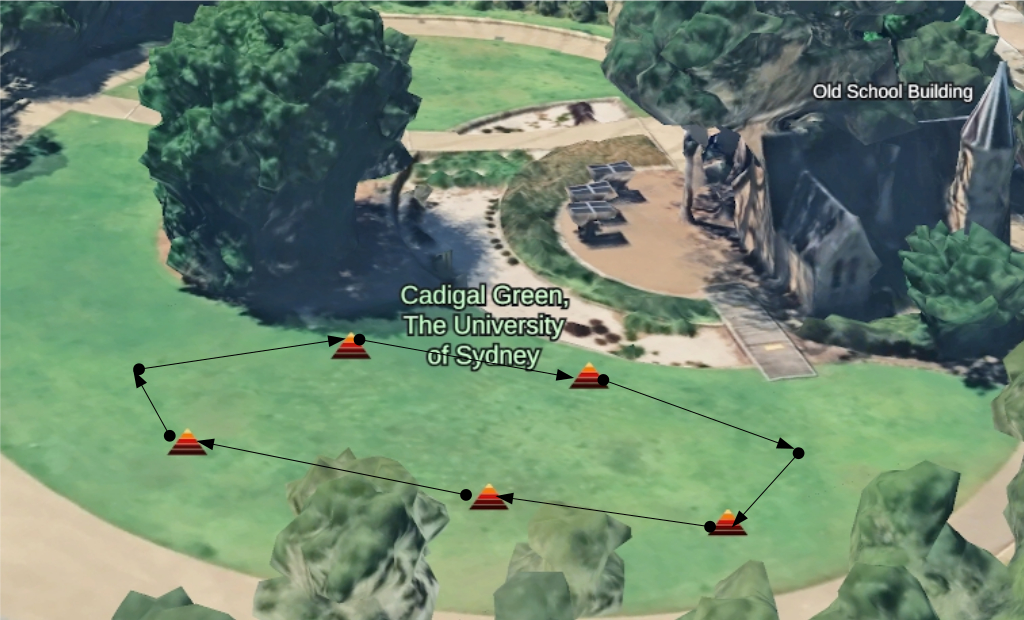} &
		\includegraphics[width=0.4\linewidth]{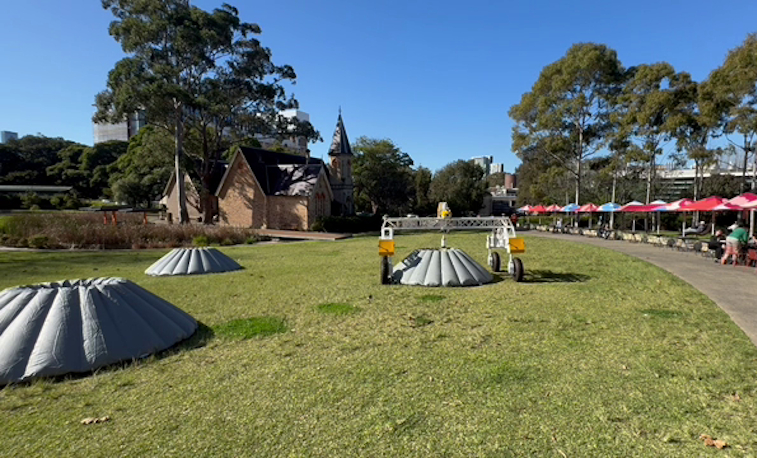} \\
		(a) 5 Hole Blast Pattern at Usyd Campus &
		(b) Dipping on an inflatable cone\\
	\end{tabular}	
	\caption{Mission Test: 5 Hole blast pattern at Usyd campus}
	\label{fig:usyd-campus-test}
\end{figure}
We performed navigation system and mission control test at Sydney University Campus. The test involves running an inspection mission on a blast pattern of five holes. We prepared bespoken inflatable cones to emulate the LIDAR profile of an average blast-hole cone, each with a 40cm hole in the centre. The cones are arranged as a two column arrangement blast pattern at the Usyd Campus, each purposefully positioned with a meter grade offset error to some marked GPS point. This is shown in Figure \ref{fig:usyd-campus-test} below. DIPPeR is able to repetitively complete the entire mission of seeking and dipping every hole, without manual intervention. For each hole, DIPPeR acted according to the state machine, by first seeking the pre-recorded GPS point up to a rough position, then switching to motion planning to target stage after detection the target, and at close distance was able to servo to the hole to dip its sonde.

\subsection{Comparison to Learning-based Method -- on Real-world Data}
\begin{figure}[h]
	\centering
	\includegraphics[width=0.92\linewidth]{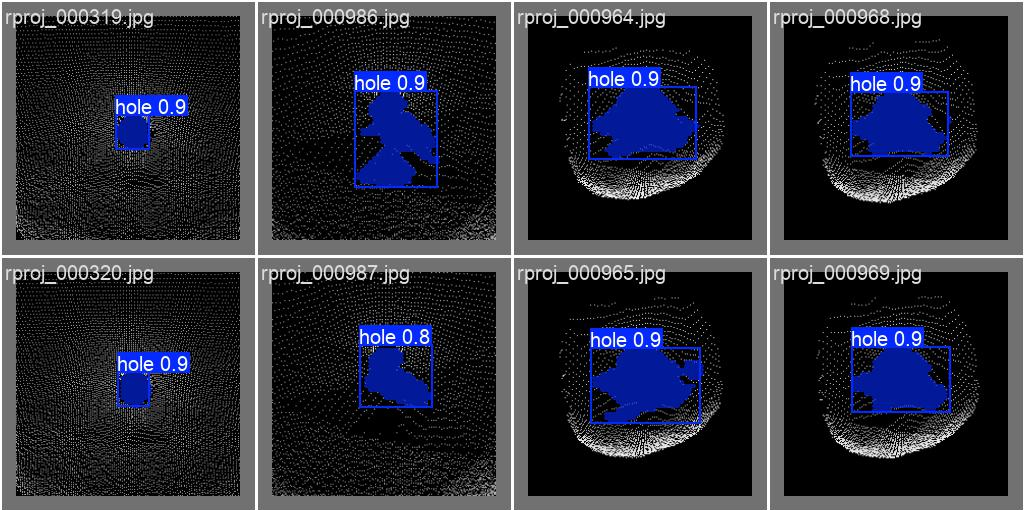}
	\caption{\small{Yolo 11 segmentation using our detection outputs for training.}}	
	\label{fig:yolo-results}
\end{figure}

Despite not taking a learning-based approach in our work, we did explore the potential of training a detection network on the classifications generated by our classical detection pipeline. 
We auto-generated 916 pairs of images and marked hole regions, -- similar to examples in Figure \ref{fig:dist-vs-camera}(c-e) -- to train Yolo 11 \cite{yolo11} for hole segmentation. We used 711 pairs as training set, 205 pairs for validation. Majority of the segmentation results match our expectation for coarse stage projection.
A selection of training validation results are shown in Figure \ref{fig:yolo-results}. 

\begin{figure}[h]
	\centering	
	\setlength\tabcolsep{1pt}	
	\begin{tabular}{|c|c|c|c|c|}
		\hline
		Ctgry	& \small{Generic} & \small{Feature} & \small{Phantom}  & \small{Hole centre}\\
		\hline
		\multirow{2}{*}{Ours}
		&
		\includegraphics[width=0.20\linewidth]{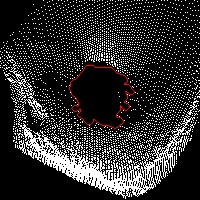}
		&
		\includegraphics[width=0.20\linewidth]{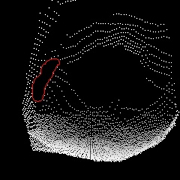}
		&
		\includegraphics[width=0.20\linewidth]{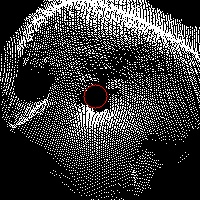}	
		& 		
		\includegraphics[width=0.20\linewidth]{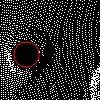}
		\\
		\cline{2-5}
		&\small{201 out of 240} & \small{2 out of 38} & \small{88 out of 109} & \small{80 out of 92}\\
		\hline
		\multirow{2}{*}{Yolo 11}
		&
		\includegraphics[width=0.20\linewidth]{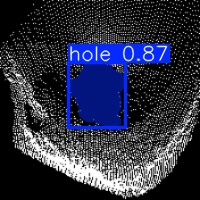}
		&
		\includegraphics[width=0.20\linewidth]{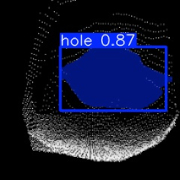}
		&
		\includegraphics[width=0.20\linewidth]{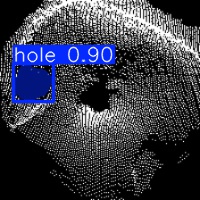} 
		& 	
		\includegraphics[width=0.20\linewidth]{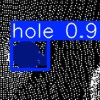}
		\\
		\cline{2-5}
		&\small{220 out of 240} & \small{36 out of 38} & \small{10 out of 109} & \small{20 out of 92}\\
		\hline
	\end{tabular}	
	\caption{Yolo segmentation comparison in 4 categories, success counted out of test size. Yolo excels at identifying off-nominal hole profiles (column 2), but performs poorly in rejecting phantom holes (column 3), and give less accurate hole centre estimate (column 4).  }	
	\label{fig:yolo-good}
\end{figure}

We also performed comparison for four specific categories of tests. The results are shown in 
Figure \ref{fig:yolo-good}. For the generic category (column 1, 240 samples) the two approaches have comparable success rate, indicating the potential for the classical pipeline's use for self-supervised learning. For distant cone scenario (column 2), Yolo outperforms our method in identifying large holes, due to its richer feature-space.
However, in latter detection stages, our scheme demonstrates superior performance in terms of phantom hole rejection (column 3) and hole centre accuracy (column 4), thanks to our Non-Maximum Suppression and RANSAC-based Least Squares circle fitting methods. The latter can be explained as follows: CNN features, analogous to the HOG features, are best for data with an underlying grid-like or Euclidean structure where the extraction operation takes the form of shifting a filter in  the grid space along the direction of its spanning axis \cite{bronstein-geometric}. The resultant feature maps have a polygonal shape with piece-wise linear edges. Circles, owing to their  curved geometry require a rotation invariant detection algorithm.

Overall our classic approach is more practical for this application without requiring data collection and labelling. Moreover, it is easy to reconfigure when hole size specification changes.

\subsection{Long Duration Test -- Simulation}
\begin{figure}[h]	
	\centering
	\includegraphics[width=0.95\linewidth]{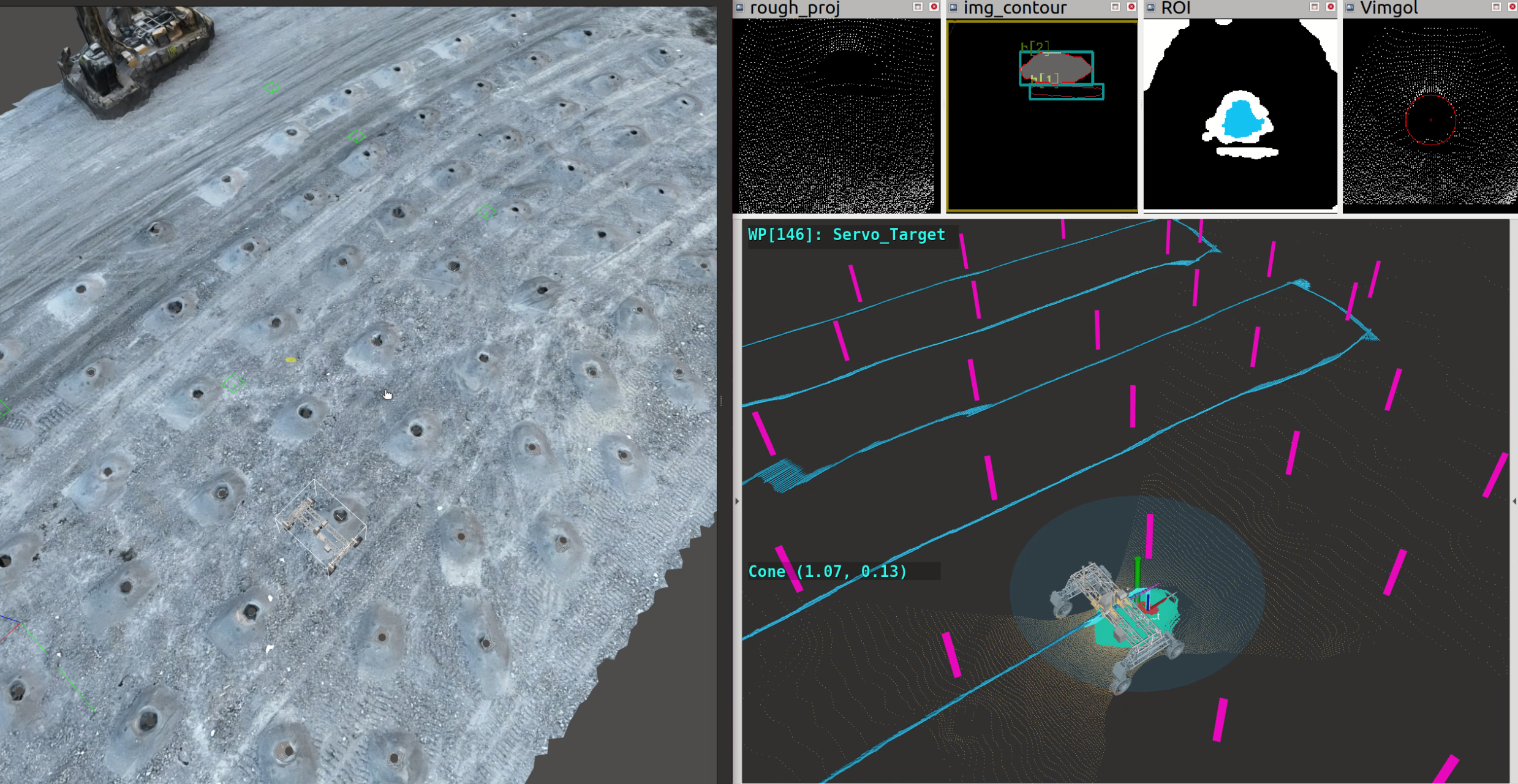}
	\caption{Extended mission test in Gazebo simulation.  \\
		\textbf{\textcopyright Rio Tinto 2025, All Rights Reserved. 3D surface model generated using  DroneDeploy, Inc. software. Visualised using RViz[Noetic/Open-Robotics] and Gazebo $[$11.0/Open-Robotics$]$.} }
	\label{fig:cm-5}
\end{figure}

We simulated our autonomous mission system in the Gazebo environment, shown in Figure \ref{fig:cm-5}. The terrain mesh model was generated by  DroneDeploy\texttrademark \cite{web-drone-deploy}. We further edited the mesh to add hole cylinders. Mission plans were created for this world including hole GPS positions for inspection. Our system can successfully complete a test 58-hole mission repetitively.

\subsection{Site-Trials}
We conducted site trial run of DIPPeR in Western Australia in 2023 and Utah in 2024 in two different mines that produced different minerals with different cone geometries and hole diameters. The WA trial was conducted during the hotter months, when temperatures inside the robot's sealed metal enclosures can reach up to 100°C. The Utah trial was conducted in sub-zero conditions during snowfall, with the ground covered in ice and snow. 

When trial runs are conducted during active production, the experiments are brief because 
the time available between drilling and charging for blasting may be limited to a few hours.
As a result, access to any given blast pattern is extremely time-limited. During these trials, the onboard sensing and perception system operated in real time, while all sensor data was recorded for offline analysis and algorithm refinement. This approach enabled us to use real-world data to validate improvements made between trials, helping us prepare more effectively for each subsequent run.

\begin{figure}[h]
	\centering
	\includegraphics[width=0.98\linewidth]{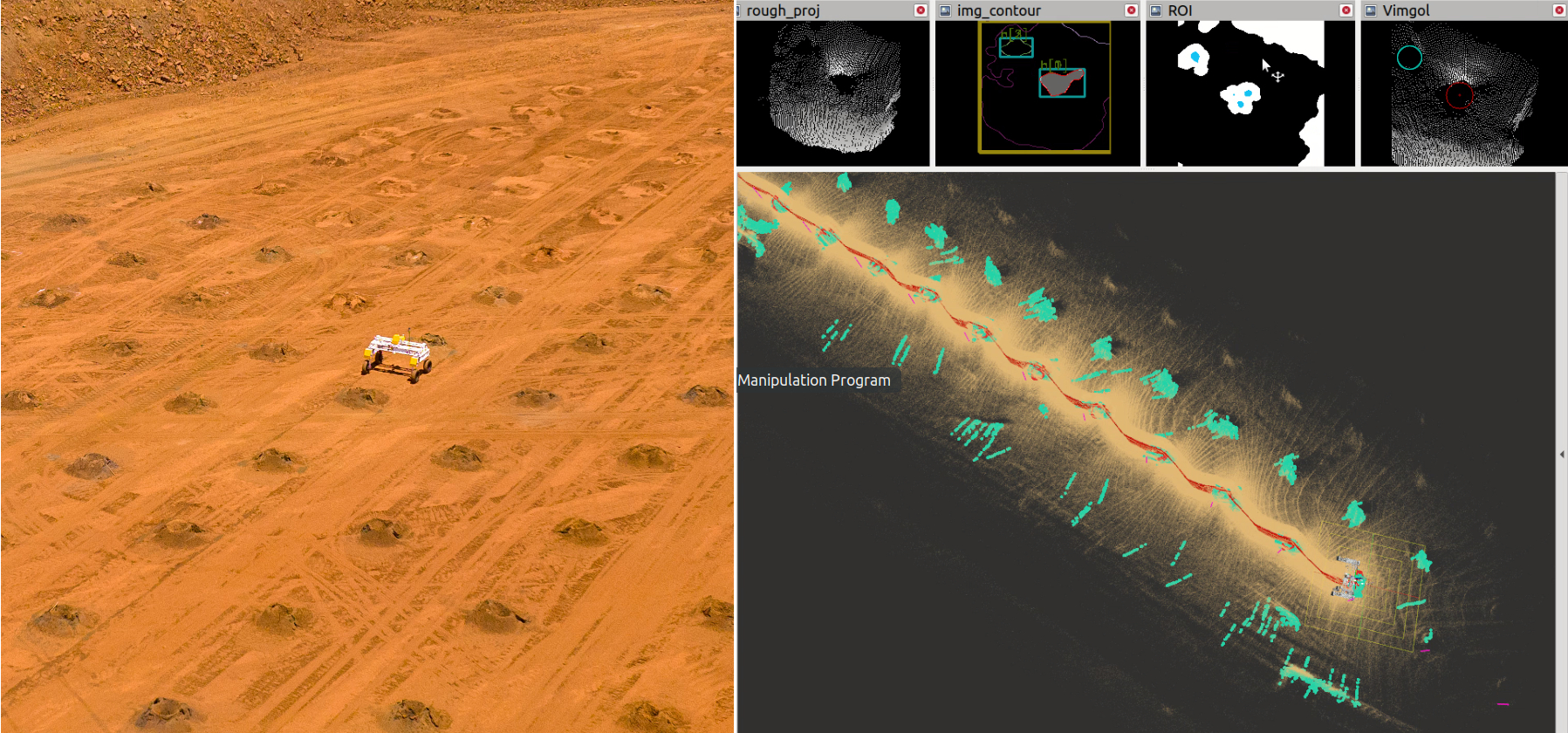}
	\caption{Long Duraton Test at WA, Australia.\\ \textbf{\textcopyright Rio Tinto 2025, All Right Reserved. Inspection visualised using RViz[Noetic/Open-Robotics].}}
	\label{fig:hd4-last-day}
\end{figure}

For the WA trial run (Figure \ref{fig:hd4-last-day}), the cones were well-formed, 95 holes were successfully detected, sensor to hole alignment was perfect (successful nearly 100\% of the attempts) for the 38 holes dipped. However, in the Utah trial run the shape of the drill waste resembles a bowl rather than a cone, causing severe occlusion in current LiDAR FoV. Detection and alignment were successful more than half of the attempts out of 70 inspected holes. 

The success rate achieved in both trial runs—particularly in the former environment—has been promising. It demonstrated the effectiveness of the proposed concept for efficient and informative hole inspection.

\section{Conclusions}

In this paper we presented a perception and navigation strategy for autonomous blast hole inspection, implemented on the DIPPeR mine site inspection robot. Effective hole inspection is demonstrated without requiring mapping, heavy-duty detection algorithms, or extensive data collection, labelling and training. 
An adaptive proximity-based navigation strategy handles the transition from imprecise GPS position to relative localisation to the blast holes, using the proposed detection algorithm for proximity navigation and servoing.
Simulations and early field testing show that our framework can track targets reliably and achieve precise positioning of the sonde above the bore holes. For sites with irregular cone geometry, more investigation is needed.

\section*{Acknowledgments}
\vspace{-7pt}
This work was supported by the Rio Tinto Sydney In	novation Hub (RTSIH) and Australian Centre for Robotics
(ACFR) at the University of Sydney. We acknowledge the
contribution of David Spray, Toniy Cimino, Phillip Gunn,
Najmeh Kamyabpour, Jerome Justin, Mehala Balamurali and
Tim Bailey to the project.
	
\clearpage
\bibliographystyle{unsrtnat}
\bibliography{blast_hole_ax_simple}  

\clearpage
\appendix
\section*{Appendices}
\renewcommand{\thesubsection}{\Alph{subsection}}

\subsection{FRST Feature Extraction} 
\label{app:FRST-extraction}
A simple Sobel fitler is applied to the output of the morphological operation and a gradient image is produced, showing clear edges of the virtual image. FRST extraction and circle fitting all take place in this gradient image.

To quickly identify circular candidates,  we perform Fast Radial Symmetry Transform (FRST) \cite{loy-et-al} on the depth image to quickly identify circle features. This gives a rough estimate of centres of potential circles. We then use these candidates as starting points to refine the centre position and search for radius. We give a summary of FRST calculation here for the sake of completeness. 

The FRST maintains an orientation projection image $O_n$ and a magnitude projection image $M_n$ for each possible circle radius $n$. These images are generated by examining the gradient $g$ at each point $\mathbf{p}$. The gradient vector points to two potential affected pixels, both at $n$ distance away: a positively-affected pixel
$\mathbf{p}_{+ve}(\mathbf{p})$ and a negatively-affected pixel $\mathbf{p}_{-ve}(\mathbf{p})$. The coordinates are given by
\begin{equation}
\begin{aligned}
\mathbf{p}_{+ve}(\mathbf{p}) =& \mathbf{p} + \text{round}(\frac{g(\mathbf{p})}{|\mathbf{p}|}n)\\
\mathbf{p}_{-ve}(\mathbf{p}) =& \mathbf{p} - \text{round}(\frac{g(\mathbf{p})}{|\mathbf{p}|}n)
\end{aligned}
\end{equation}

With this, we can accumulate the orientation and projection counts stored in the $O$ and $M$ images. For each pair of affected pixels, the counters are incremented (\ref{eq:frst-inc}):

\begin{equation}
\begin{aligned}
O_n(\mathbf{p}_{+ve}(\mathbf{p})) =& O_n(\mathbf{p}_{+ve}(\mathbf{p})) + 1\\
O_n(\mathbf{p}_{-ve}(\mathbf{p})) =& O_n(\mathbf{p}_{-ve}(\mathbf{p})) - 1\\
M_n(\mathbf{p}_{+ve}(\mathbf{p})) =& M_n(\mathbf{p}_{+ve}(\mathbf{p})) + |g(\mathbf{p})|\\
M_n(\mathbf{p}_{-ve}(\mathbf{p})) =& M_n(\mathbf{p}_{-ve}(\mathbf{p})) - |g(\mathbf{p})|\\
\end{aligned}
\label{eq:frst-inc}
\end{equation}
The radial symmetry contribution at a range n is defined as the convolution
\begin{equation*}
S_n = F_n * A_n
\end{equation*}
where
\begin{equation*}
\begin{aligned}
F_n(\mathbf{p}) &= |O_n(\mathbf{p})|^{(\alpha)} M_n(\mathbf{p}),\\
\tilde{O}_n(\mathbf{p}) &= \frac{|O_n}{ max_{\mathbf{p}}\{|O_n(\mathbf{p})|\}},\\
\tilde{M}_n(\mathbf{p}) &= \frac{|M_n}{ max_{\mathbf{p}}\{|M_n(\mathbf{p})|\}}.\\
\end{aligned}
\end{equation*}
Here, $\alpha$ is the radial strictness parameter, and $A_n$ is a two-dimensional Gaussian.

The magnitude image $M_n$ is essentially a 2-D histogram of possible locations of circle centres. One can smooth this image and locate peaks which serve as candidates for circle centres.
We extract from the gradient image points that are within valid range to the candidate centres and label them as Region of Interest(ROI).

\subsection{Taubin's Circle Fitting Algorithm -- conversion from Constrained Optimisation to Unconstrained Optimisation} 
\label{app:taubin}

Given a set of image points, Taubin's Least Squares (LS) circle fit method \cite{taubin:circle-fit} allows us to compute the optimal circle centre and radius. It is essentially a constrained optimisation problem and can be reduced to an Eigen value problem \cite{taubin-error-analysis}.  We now continue from the objective function arrived at Equation (\ref{eqn:Taubin}) in Section \ref{sec:ransac-circle-fit}. 

Represent the algebraic fits in Matrix form:
\begin{equation}
\begin{aligned}
\text{Let } & \mathbf{A} = (A,B,C,D) \text{ denote the parameter vector}\\
\text{ Define } & \text{data matrix: }\\
& \mathbf{Z} = \begin{bmatrix} z_1 & x_1 & y_1 & 1 \\
. & . & . & .\\
. & . & . & .\\
. & . &. & .\\
z_n & x_n & y_n & 1
\end{bmatrix}\\
\text{then, } & \text{the matrix of moments is }\\
& \mathbf{M} = \frac{1}{n}\mathbf{Z}^{\intercal} \mathbf{Z} = \begin{bmatrix}
\overline{zz} & \overline{zx} & \overline{zy} & \overline{z} \\
\overline{zx} & \overline{xx} & \overline{xy} & \overline{x} \\
\overline{zy} & \overline{xy} & \overline{yy} & \overline{y} \\
\overline{z} & \overline{x} & \overline{y} & 1 \\
\end{bmatrix} \\
\end{aligned}
\end{equation}

The objective function can be rewritten as: 
\begin{equation}
\begin{aligned}
& \mathcal{F}(\mathbf{A}) = \mathbf{A}^{\intercal}\mathbf{M} \mathbf{A},\\
\text{subject to: } & \mathbf{A}^{\intercal}\mathbf{N} \mathbf{A} = 1\\
\text{where } &\mathbf{N} = \begin{bmatrix}
4\bar{z} & 2\bar{x} & 2\bar{y}  & 0 \\
2\bar{x} & 1 & 0  & 0 \\
2\bar{y} & 0 & 1  & 0 \\
0 & 0 & 0  & 0 \\
\end{bmatrix}
\end{aligned}
\end{equation}
This constrained minimization problem can be reduced to  an unconstrained minimization of the function 
\begin{equation}
\begin{aligned}
\mathcal{G}( \mathbf{A}, \eta ) &= \mathbf{A}^\mathbf{M}\mathbf{A} - \eta (\mathbf{A}^{\intercal}\mathbf{N}\mathbf{A} - 1); \\
\text{Differentiating} & \text{ with respect to } \mathbf{A} \\
\mathbf{M}\mathbf{A} &= \eta \mathbf{N}\mathbf{A}
\end{aligned}
\end{equation}

Therefore $\mathbf{A}$ must be a generalized Eigen vector for the matrix pair $(\mathbf{M}, \mathbf{N})$.

\end{document}